\documentclass[lettersize,journal]{IEEEtran}
\usepackage{amsmath,amsfonts}
\usepackage{algorithmic}
\usepackage{algorithm}
\usepackage{array}
\usepackage[caption=false,font=normalsize,labelfont={sf, footnotesize},textfont={sf, footnotesize}]{subfig}
\usepackage{textcomp}
\usepackage{stfloats}
\usepackage{url}
\usepackage{verbatim}
\usepackage{graphicx}
\usepackage{cite}
\usepackage{setspace}
\usepackage{multirow}

\usepackage{tikz,xcolor,hyperref}
\definecolor{lime}{HTML}{A6CE39}
\DeclareRobustCommand{\orcidicon}{%
    \begin{tikzpicture}
    \draw[lime, fill=lime] (0,0) 
    circle [radius=0.16] 
    node[white] {{\fontfamily{qag}\selectfont \tiny ID}};    \draw[white, fill=white] (-0.0625,0.095) 
    circle [radius=0.007];    \end{tikzpicture}
    \hspace{-2mm}}
\foreach \x in {A, ..., Z}{%
    \expandafter\xdef\csname orcid\x\endcsname{\noexpand\href{https://orcid.org/\csname orcidauthor\x\endcsname}{\noexpand\orcidicon}}
    }

\begin{document}

\title{Noise and Edge Based Dual Branch Image Manipulation Detection}

\author{Zhongyuan Zhang\orcidA{}, Yi Qian{}, Yanxiang Zhao\orcidB{}, Lin Zhu{}, and Jinjin Wang
\thanks{Zhongyuan Zhang, Yi Qian, and Yanxiang Zhao are with the Department of Computer Science and Technology, Xi'an Jiaotong University, Xi'an 710061, China (e-mail: zhangzhongyuan@stu.xjtu.edu.cn; yqian@xjtu.edu.cn; yanxiangzhao@stu.xjtu.edu.cn).

Lin Zhu and Jinjin Wang are with the China Mobile Research Institute, Beijing 100032, China (e-mail: zhulinyj@chinamobile.com; wangjinjin@chinamobile.com).}}



\maketitle

\begin{abstract}
Unlike ordinary computer vision tasks that focus more on the semantic content of images, the image manipulation detection task pays more attention to the subtle information of image manipulation. In this paper, the noise image extracted by the improved constrained convolution is used as the input of the model instead of the original image to obtain more subtle traces of manipulation. Meanwhile, the dual-branch network, consisting of a high-resolution branch and a context branch, is used to capture the traces of artifacts as much as possible. In general, most manipulation leaves manipulation artifacts on the manipulation edge. A specially designed manipulation edge detection module is constructed based on the dual-branch network to identify these artifacts better. The correlation between pixels in an image is closely related to their distance. The farther the two pixels are, the weaker the correlation. We add a distance factor to the self-attention module to better describe the correlation between pixels. Experimental results on four publicly available image manipulation datasets demonstrate the effectiveness of our model. 
\end{abstract}


\section{Introduction}
\IEEEPARstart{W}{ith} the development of technology, people can easily use image editing software such as Photoshop and GIMP to obtain a manipulated image. It is often difficult for human eyes to discern traces of manipulation. While image editing technology brings convenience, it also brings some problems. If it is used for unlawful purposes, it can have serious negative consequences. To solve this problem, a general image manipulation detection technology is needed.

Image manipulation detection is different from general computer vision (CV) tasks, such as object detection and semantic segmentation. Ordinary CV tasks focus on learning the semantic content of images, while image manipulation detection pays more attention to the details left by manipulation operations. Traditional image manipulation detection uses methods such as image color consistency \cite{6522874}, photo response non-uniformity \cite{6719568}, keypoint similarity of image blocks \cite{6987281}, and blur type inconsistency \cite{7015579} to identify the difference between manipulation and non-manipulation regions. Nevertheless, these methods can only detect a specific type of manipulation and are not very general. In addition, they cannot provide accurate pixel-level detection results.

Recent studies on image manipulation detection are mainly based on convolutional neural networks (CNN). As the network depth increases, existing CNN architectures tend to learn features representing the semantic content of images, which is not entirely consistent with the purpose of the image manipulation detection task. To solve this problem, the image is processed by the spatial-domain rich model (SRM) \cite{6197267} or constrained convolution \cite{8335799} to obtain the corresponding noise image. Then the noise image is input into the subsequent neural network. Nevertheless, the weights of SRM are fixed and cannot be well adapted to different datasets. Due to the constrained process, the weights of constrained convolutions are prone to drastic changes in practical training (detailed explanation in Section \ref{subsec_cc}). So, we propose an improved constrained convolution to obtain the corresponding noise image.

\IEEEpubidadjcol
The manipulation edge is critical information for the manipulation detection task. Because the three basic manipulation operations of copy-move, splicing, and removal will leave artifacts on the manipulation boundary. To capture such subtle manipulation information, the features in the CNN need to maintain high resolution. Due to the local characteristics of CNN, a certain degree of downsampling is indispensable to obtain sufficient image context information. So, both the bilateral segmentation network (BiSeNet) \cite{Yu_2018_ECCV} and the deep dual-resolution network (DDRNet) \cite{hong2021deep} use dual branches to construct the model. One of the branches is used to obtain the context information of the image, and the other branch maintains a certain resolution to avoid losing too much detail. We build a model based on this dual branch structure and design an edge prediction module to detect manipulation edges from the features of the dual branch. With manipulation edge supervision, the overall detection performance improves a lot.

Global correlation information is necessary for manipulation detection. Still, due to the local characteristics of CNN, it is difficult for CNN-based models to obtain global correlations. So, similar to self-attention \cite{vaswani2017attention} in the field of natural language processing (NLP), the non-local module \cite{wang2018non} is used in CNN to obtain the correlations between each pixel and all other pixels, i.e. the global correlation. However, unlike NLP, the spatial correlations among image pixels are more affluent because the spatial correlations among words in NLP are one-dimensional. In contrast, the spatial correlations among pixels are two-dimensional. The non-local module can break the distance limit to calculate the correlation between pixels but also ignore their distance relationship simultaneously. Therefore, we add a distance metric to the non-local module so that it can better describe the global correlations among pixels.

Overall, the main contributions of this work are as follows.
\begin{itemize}
\item{We optimize the constrained convolution process to make constrained convolution more stable and easier to train.}
\item{We design a dual branch network and the corresponding manipulation edge detection module. The overall manipulation detection effect is improved by edge detection.}
\item{We add the distance factor to the non-local module so that the non-local module can better capture the global correlations among pixels.}
\item{The results of four publicly available manipulation datasets demonstrate that our model has advantages over state-of-the-art manipulation detection models.}
\end{itemize}

\section{Relate Work}\label{relate_work}
Image manipulation includes three primary methods: copy-move, splicing, and removal. Some works would identify a specific manipulation method. Wu et al. \cite{Wu_2018_ECCV} designed a parallel dual branch network consisting of a manipulation detection branch and a similarity detection branch to detect copy-move type manipulation. The network can distinguish the source and the target regions. Islam et al. \cite{9157762} proposed a generative adversarial network based on a two-order attention mechanism to detect and localize copy-move manipulation. Mayer et al. \cite{8744262} extracted the features of two blocks of an image through CNN and calculated the similarity of the blocks to judge whether the two blocks have different forensic traces. The forensic trace is irrelevant to semantic content. If the forensic traces are different, there is a splicing operation.

There are certain limitations to the generality of the above works. Although the three types of manipulation are pretty different, there are commonalities in the traces of manipulation they leave behind. As shown in Table~\ref{tab_similarity}, generally speaking, the manipulation region and the non- manipulation region of copy-move type manipulation come from the same image, so the two are similar in the original image and the noise image. The manipulation region and the non- manipulation region of splicing type manipulation come from different images, so the two are not similar in the original image and the noise image. Since there are many methods in the removal type manipulation, the manipulation region and the non- manipulation region may be similar or dissimilar in the original image and the noise image. Additionally, all three types of manipulation leave inconsistent traces at the manipulation edge. The inconsistency in the original image (semantic content) cannot indicate manipulation. And the inconsistency in noise or edge can be used as proof of manipulation. So, a generic image manipulation detection can be achieved by utilizing the image’s noise and edge information.

\begin{table}
\centering
\caption{The similarity between the manipulation area and the non- manipulation area of the three manipulation methods in different fields. Y means the two are similar, and N means the two are not similar}
\label{tab_similarity}
\begin{tabular}{c|c|c|c} 
\hline
~         & Original
  image & Noise & Edge  \\ 
\hline
Copy-move & Y                & Y     & N     \\ 
\hline
Splicing  & N                & N     & N     \\ 
\hline
Removal   & Y / N            & Y / N & N     \\
\hline
\end{tabular}
\end{table}

Zhou et al. \cite{8578214} used SRM to extract the noise of an image. The original image and noise are simultaneously fed into Faster R-CNN \cite{NIPS2015_14bfa6bb} for general-type manipulation detection. Since the value of SRM is fixed, Bayar et al. \cite{8335799} impose certain constraints after back-propagation of the ordinary convolution kernel, so that this convolution kernel can both train and suppress the semantic content of the image. Because Faster R-CNN is unsuitable for outputting pixel-level results, Yang et al. \cite{9102825} added constrained convolution to Mask R-CNN \cite{He_2017_ICCV} for more accurate predictions. Wu et al. \cite{Wu_2019_CVPR} used the original image and two noise images obtained by SRM and constrained convolution as the model’s input, respectively. And the long short-term memory (LSTM) cell is used to simulate the law of “near big and far small”. Ultimately, more granular manipulation operations can be identified.

Since the three basic manipulation methods will leave manipulation traces on the manipulation edge, if the manipulation edge can be detected correctly, it will benefit the overall detection effect. Salloum et al. \cite{SALLOUM2018201} added a manipulation edge detection task based on a fully convolutional network, which can simultaneously predict the region and corresponding edge of manipulation. Zhou et al. \cite{zhou2020generate} selected features from the middle three blocks of DeepLab \cite{7913730} to construct an edge detection task. And used the predicted manipulation edge to optimize the overall manipulation region prediction. Chen et al. \cite{Chen_2021_ICCV} added the traditional Sobel edge detection operator to the progressive edge extraction structure of the Border network \cite{Yu_2018_CVPR}, which can obtain better edge prediction results.

In addition, the global correlations between each pixel and other pixels are significant for the image manipulation detection task. Wang et al. \cite{wang2018non} implemented the self-attention mechanism in the CV domain similar to that in the NLP domain. Since this method is computationally intensive, Zhu et al. \cite{Zhu_2019_ICCV} reduced the computation by compressing the dimensions of key and value or using the key directly as the value. Huang et al. \cite{Huang_2019_ICCV} computed the correlations of each pixel only using the pixels at their cross-shaped positions, thus avoiding the participation of most relatively unimportant pixels. The amount of computation is reduced while maintaining the effect. The first few methods compute correlations only in the spatial dimension. Fu et al. \cite{Fu_2019_CVPR} calculated the correlations among the channel dimension and then fused them with correlations of the spatial dimension.

\section{Proposed Method}
\begin{figure*}[!t]
\centering 
\includegraphics[width=7in]{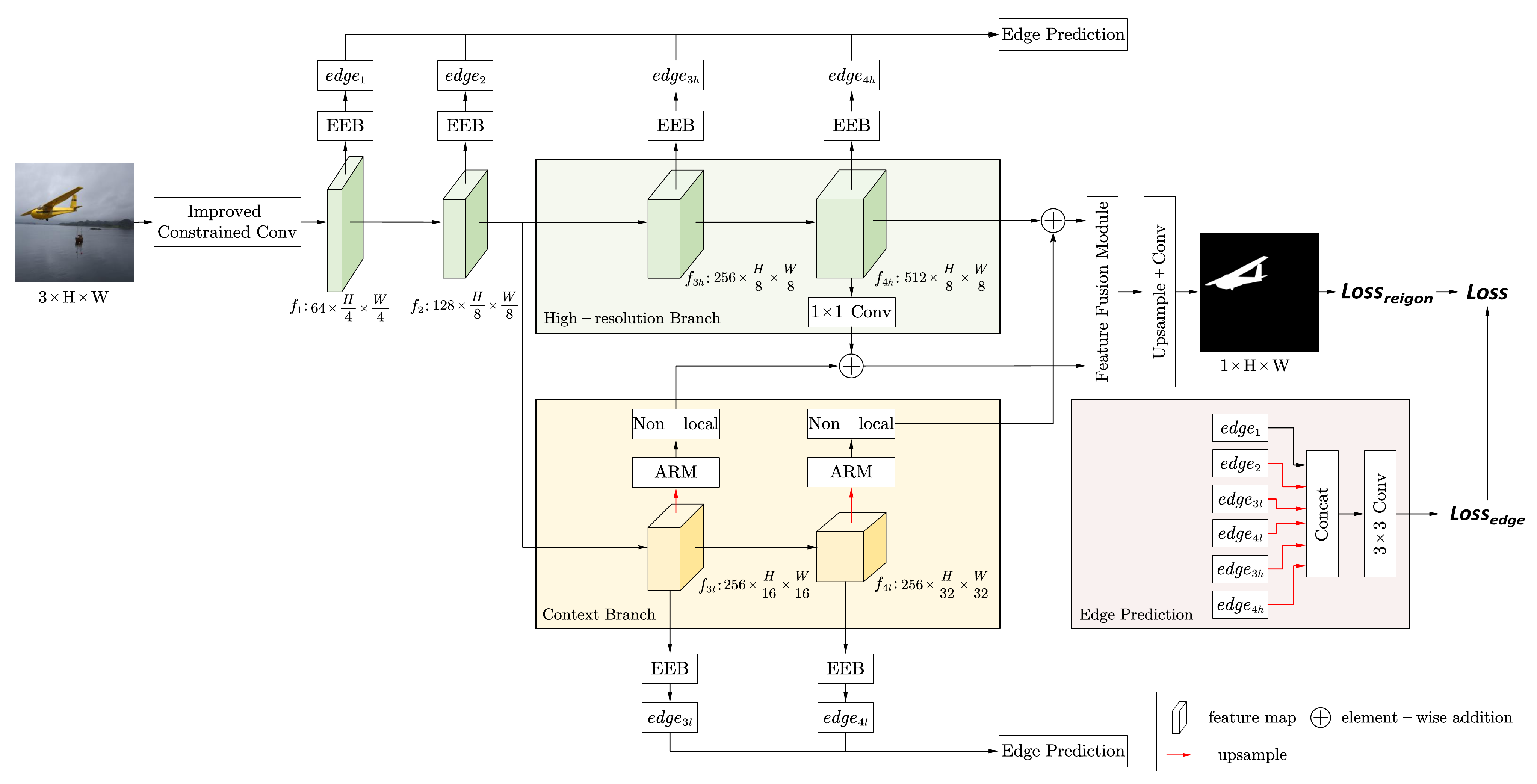}
\caption{The overall structure of the proposed model. The model consists of a high-resolution branch and a context branch. The noise image is used as the manipulation detection cue instead of the original image.}
\label{fig_model}
\end{figure*}

As shown in Fig. \ref{fig_model}, we propose a model named NEDB-Net. The improved constrained convolution processes the input image to obtain the corresponding noise image. The noise image is then fed into a dual branch network with ResNet-34 \cite{He_2016_CVPR} as the backbone. The high-resolution branch can maintain the resolution of the feature to obtain more details, and the context branch is used to obtain richer correlations among pixels. The edge extraction block (EEB) extracts corresponding manipulation edge predictions from features output from each layer in the model. And then all the edge predictions are fused to get the final prediction. Features of the context branch are optimized by the attention modules and fused with the feature of the high-resolution branch. The fused feature is upsampled and convolved to obtain the final mask of manipulation regions.

\subsection{Improved Constrained Convolution}\label{subsec_cc}
\begin{figure}[!t]
\centering
\subfloat[]{\includegraphics[width=1in]{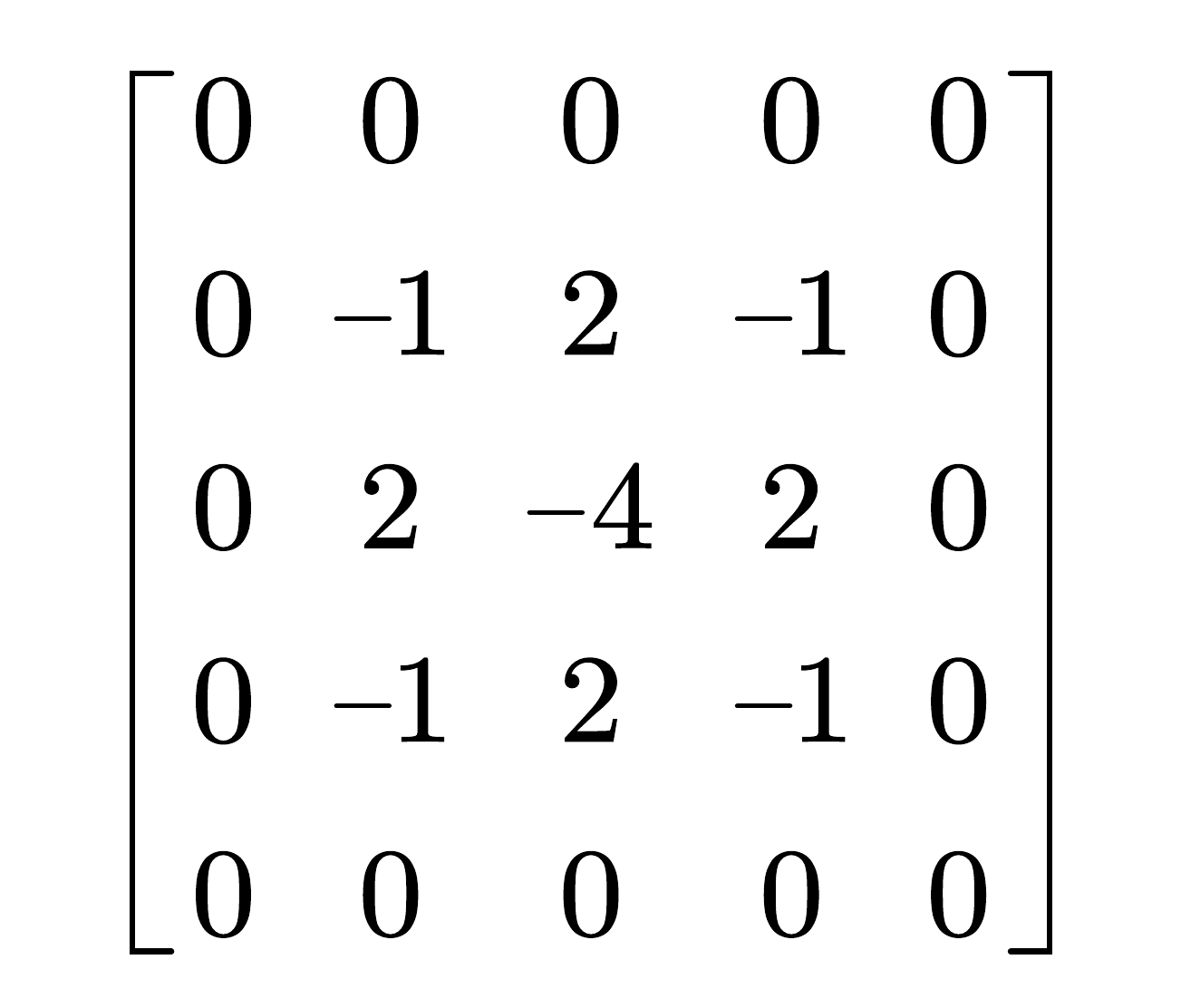}
\label{srm_a}}\hspace{1mm}
\subfloat[]{\includegraphics[width=1in]{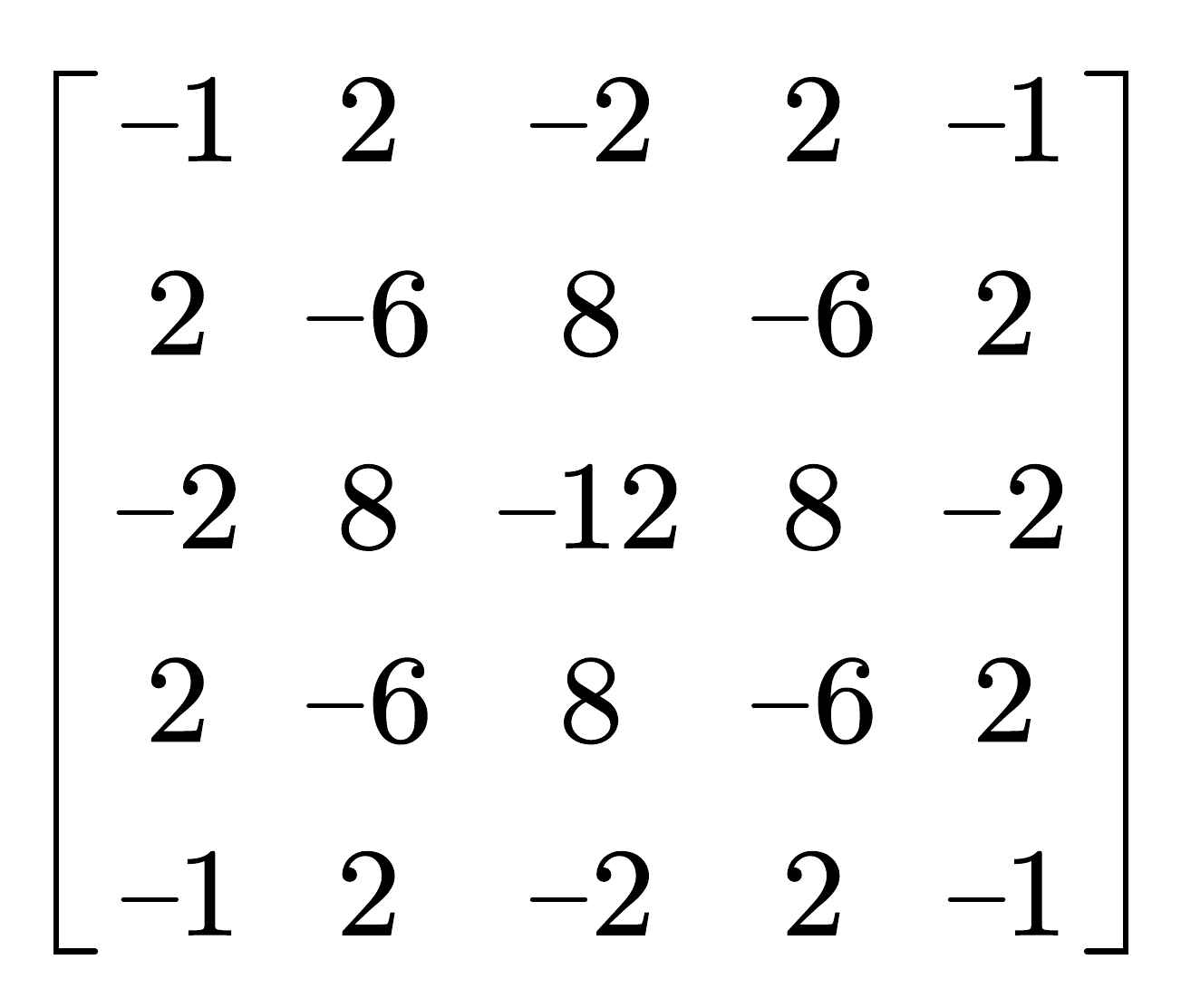}
\label{srm_b}}\hspace{1mm}
\subfloat[]{\includegraphics[width=1in]{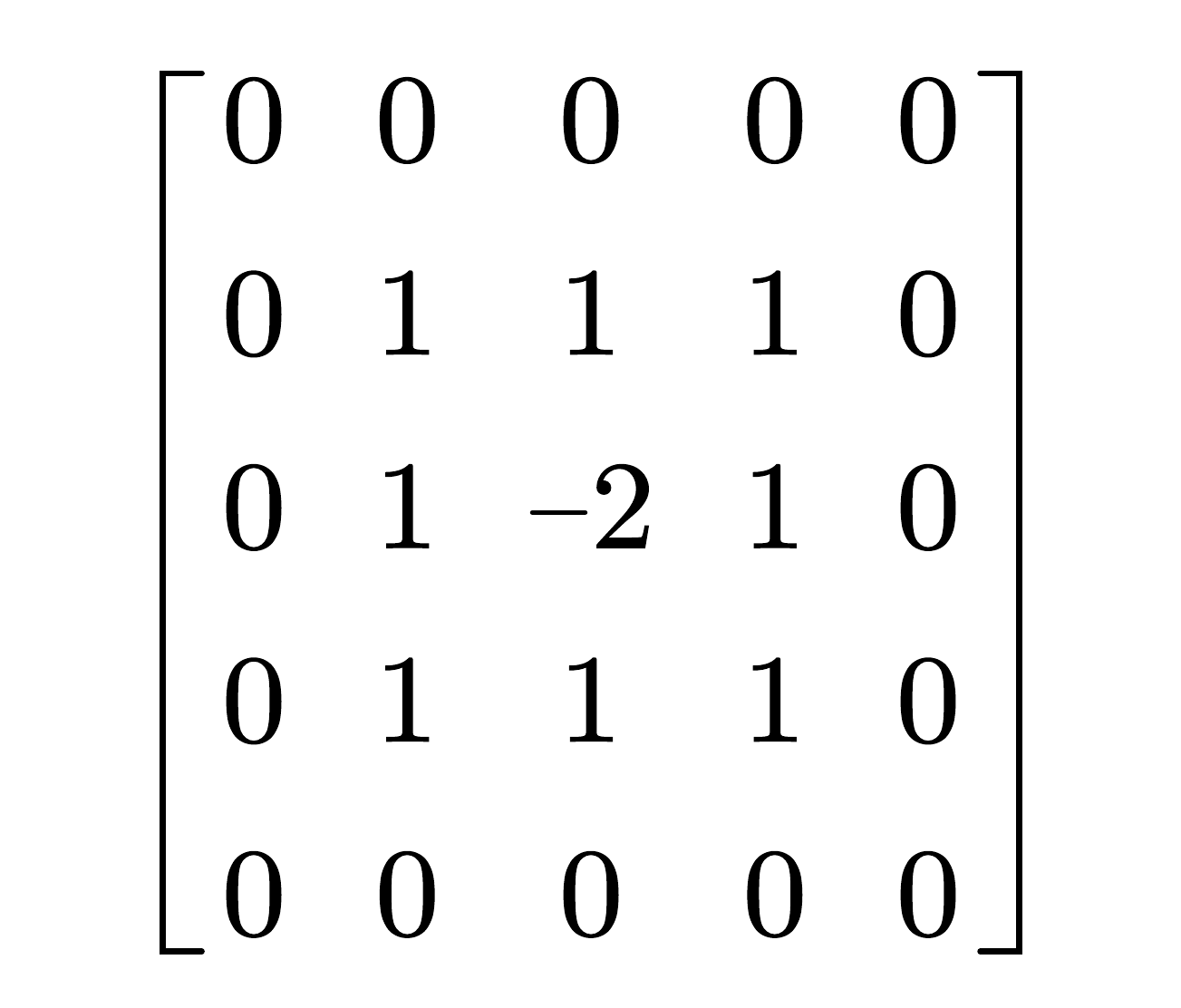}
\label{srm_c}}
\caption{Some predefined models in SRM.}
\label{fig_srm}
\end{figure}

Ordinary CNN tends to learn features representing semantic information of images rather than manipulation details. To solve this problem, some works use SRM, as shown in Fig. \ref{fig_srm}, to obtain the noise image corresponding to the image as the model input. SRM is essentially a set of high-pass filters that can get the image's different high-frequency information through different weights.

Since the weights of SRM are predefined and cannot be learned, some works use constrained convolution instead of SRM. Constrained convolution imposes constraints after kernels update weights, making the weights distribution of kernels similar to the high-pass filter. The constraints are as follows:
\begin{equation}
\left\{\begin{array}{l}
w_{k}(c,c)=-1 \\
\sum_{m, n \neq c} w_{k}(m, n)=1
\end{array}\right.,
\end{equation}
where $\boldsymbol{w}_{k}$ represents the $k$-th convolution kernel, $(c, c)$ is the center position coordinate of $\boldsymbol{w}_{k}$, and $(m, n)$ is the non-center position coordinate. After the constraints, the center position weight of the convolution kernel is -1, and the sum of the weights of other positions is 1. The constraints can be achieved by the following stpdf:
\begin{enumerate}
\item{Calculate the sum of the weights of the non-central positions.}
\item{Divide the weights of the non-center positions by the sum in 1).}
\item{Set the center position weight to -1.}
\end{enumerate}

\begin{figure}[!t]
\centering
\subfloat[]{\includegraphics[width=1in]{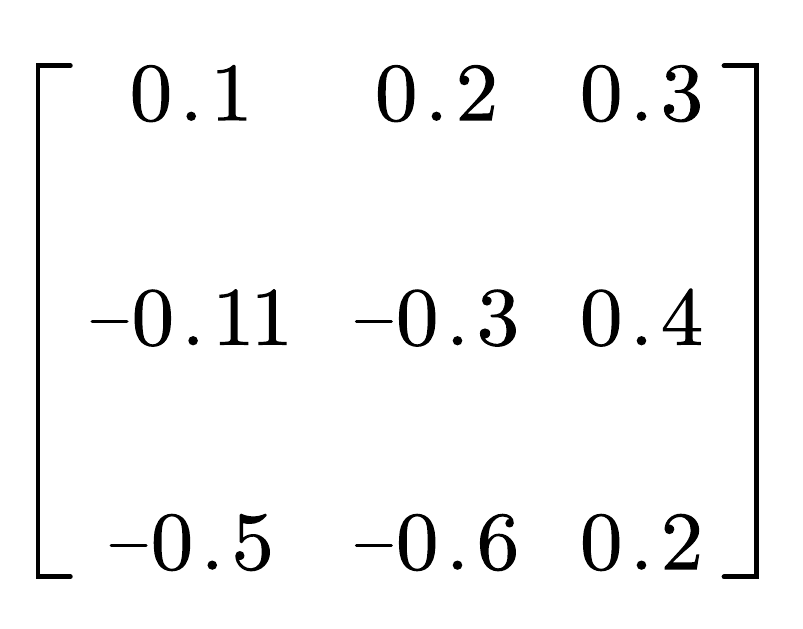}
\label{fig_cc_problem_a}}\hspace{12mm}
\subfloat[]{\includegraphics[width=1in]{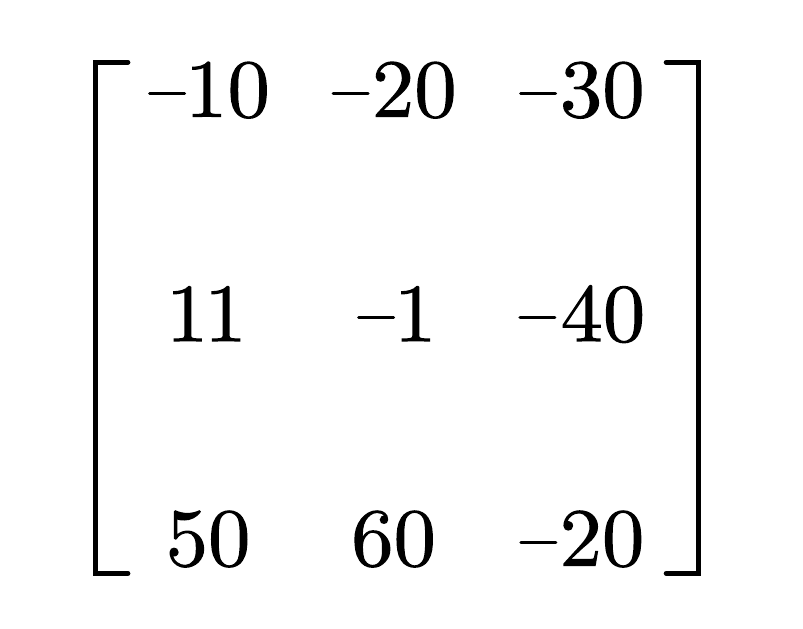}
\label{fig_cc_problem_b}}
\caption{Illustration of training fluctuations caused by constrained convolution. (a) Hypothetical convolution kernel weights and the sum of the non-center position weights is -0.01. (b) The result after the weights are constrained.}
\label{fig_cc_problem}
\end{figure}

Although constrained convolution solves the problem that SRM cannot be trained, it is unstable during actual training, resulting in poor training results. As shown in Fig. \ref{fig_cc_problem}, we assume that the $3\times 3$ matrix in Fig. \ref{fig_cc_problem}\subref{fig_cc_problem_a} is the weight of a convolution kernel updated by backpropagation. Fig. \ref{fig_cc_problem}\subref{fig_cc_problem_b} is the result after it is constrained by constrained convolution. It can be seen that since the weights before constraints may be positive or negative, their sum may be negative, and the absolute value of the sum may be relatively small. At this time, according to the constraint rules, dividing the weights of all non-central positions by this sum will cause three problems:
\begin{itemize}
\item{Dividing the weight by this relatively small sum causes the weight to be amplified a lot, but the center position weight is still -1. The weights of the non-central positions and the weight of the central position are much different in order of magnitude, which affects the effect of high-pass filtering.}
\item{If the sum is negative, the division operation will make the positive weights negative and the negative weights positive, causing the model's input to change drastically.}
\item{There is an order of magnitude difference between the weights of the constrained convolution and the weights of the subsequent normal convolution, which will affect the weights already trained and cause training fluctuations.}
\end{itemize}

\begin{figure}[!t]
\centering
\includegraphics[width=2.5in]{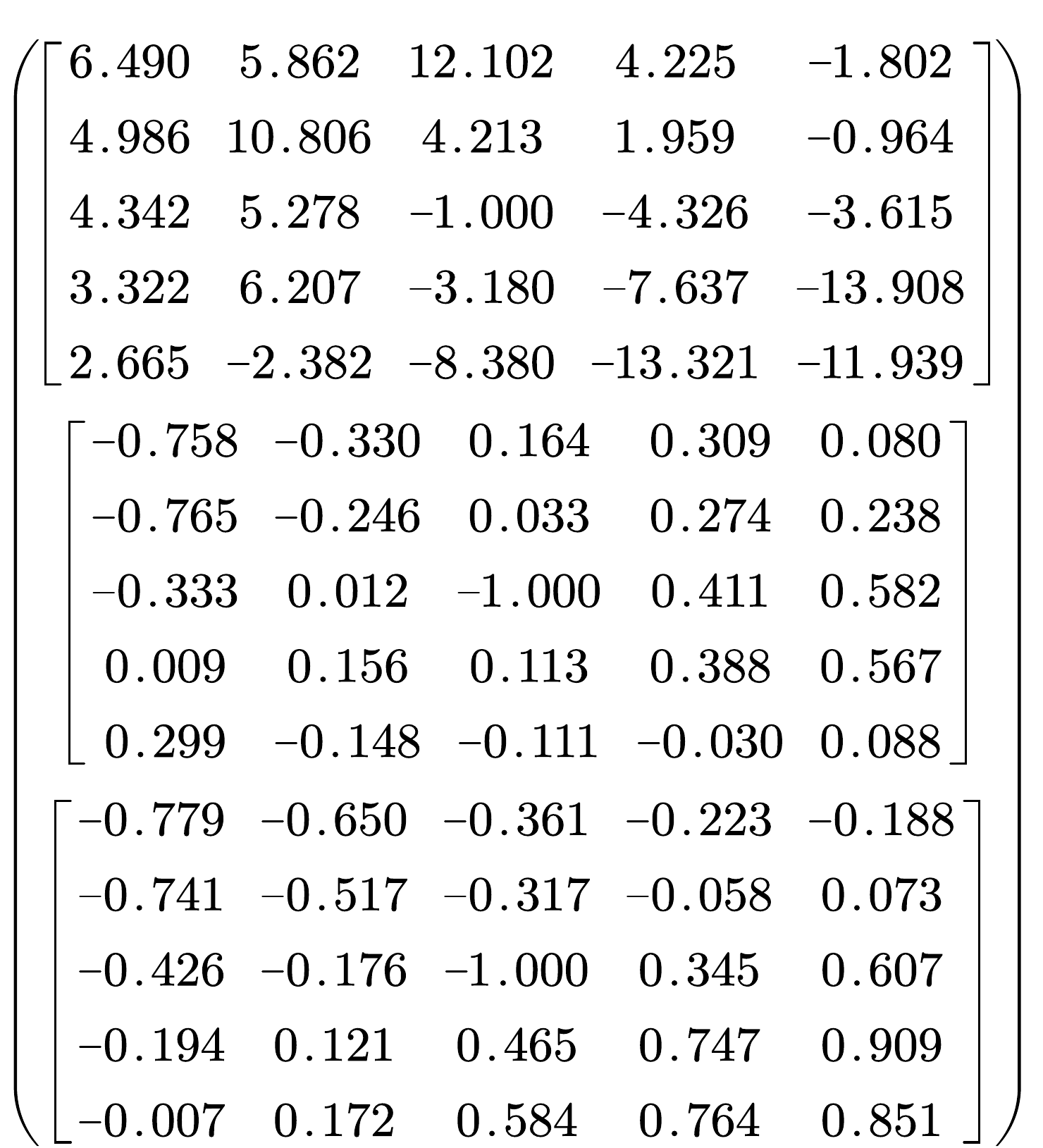}
\caption{Partial weights of constrained convolution during the actual training.}
\label{fig_cc_real}
\end{figure}

The kernel weights shown in Fig. \ref{fig_cc_real} are the result of training with constrained convolutions. It can be seen that the weights of the convolution are very different. To address the above issues, we set new constraints on convolution. The improved constrained convolution training process is shown in Algorithm 1.

\begin{algorithm}[H]
\setstretch{1.25}
    \caption{Training Algorithm for Improved Constrained Convolutional Layer}
    \begin{algorithmic}[1]
    \STATE \emph{$Initialize\ \boldsymbol{w}_{k}\text{'}s$ with the Laplacian-like weight}
    \STATE i = 1
    \WHILE{i $\le \mathit{max\_iters}$}
        \STATE \emph{Do feedforward pass}
        \STATE \emph{Update filter weights through stochastic gradient \\ descent and backpropagate errors}
        \STATE \emph{Set $\boldsymbol{w}_{k}(c, c) = 0$ for all K filters}
        \STATE \emph{Calculate the sum of the absolute values of the non-central position weights for all K filters \\ $S_k =\sum_{m, n \neq c} \left | w_{k}(m, n) \right |$}
        \STATE \emph{Let $w_{k}(m, n) = w_{k}(m, n) {\div}  S_k$ to regularize $w_{k}$}
        \STATE \emph{Set $w_{k}(m, n) = 0.001$ if $w_{k}(m, n) \leq 0.001$}
        \STATE \emph{Set $w_{k}(c, c) = -S_k$ for all K filters}
        \STATE i = i+1
    \ENDWHILE
    \end{algorithmic}
\end{algorithm}

\begin{figure}[!t]
\centering
\subfloat[]{\includegraphics[width=1in]{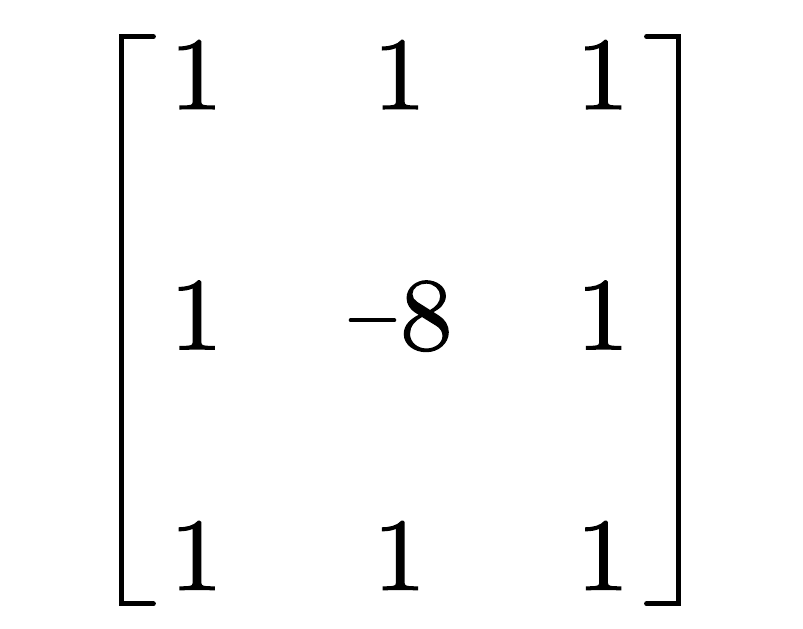}
\label{fig_laplace_operator}}\hspace{1mm}
\subfloat[]{\includegraphics[width=1in]{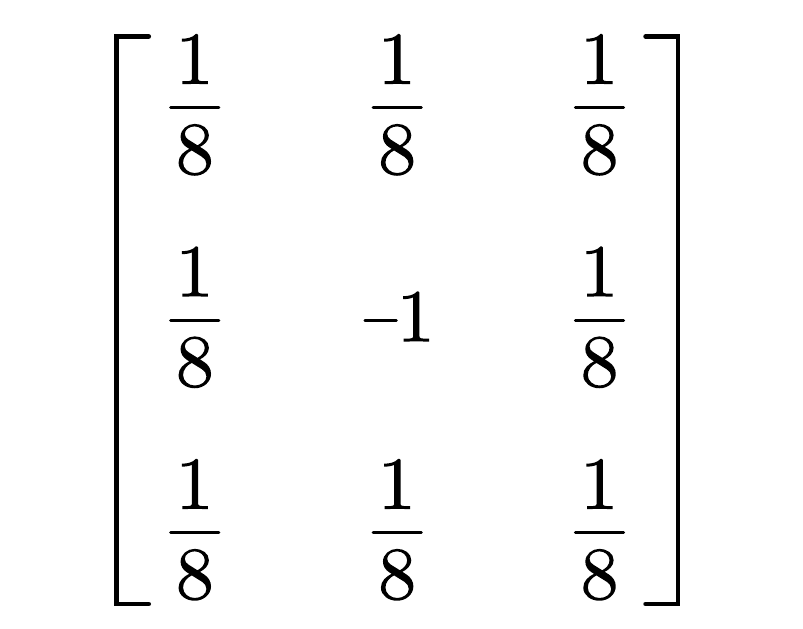}
\label{fig_cc_initial}}\hspace{1mm}
\subfloat[]{\includegraphics[width=1in]{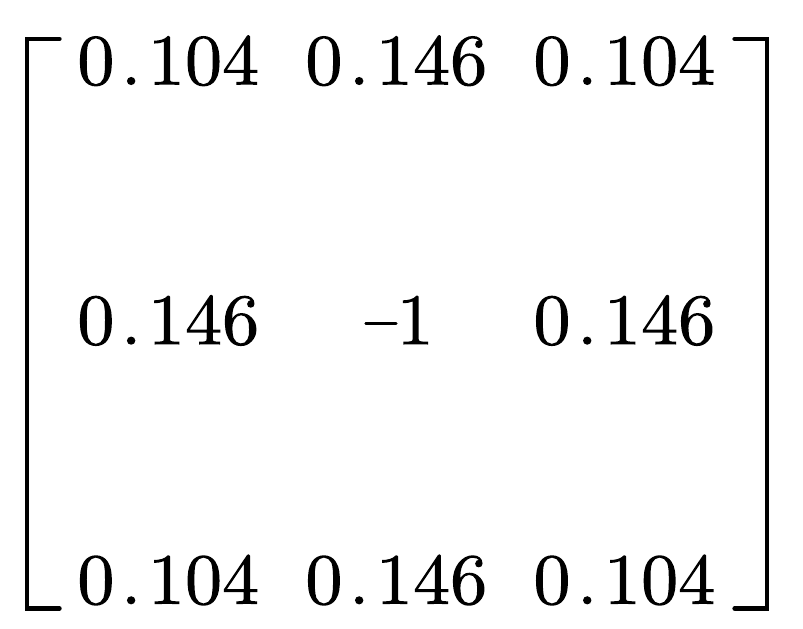}
\label{fig_cc_initial_distance}}
\caption{Constrained convolution weight initialization. (a) Laplace operator. (b) Initialize weights using the Laplace operator-like. (c) Initialize weights using the Laplace operator-like with the distance factor.}
\label{fig_cc_initials}
\end{figure}

Firstly, as shown in Fig. \ref{fig_cc_initials}\subref{fig_laplace_operator}, we learn from the traditional edge detection operator Laplacian to initialize the convolution kernel. Unlike the Laplacian operator, we set the weight of the center position to -1, and the weights of other positions equal to 1 divided by the number of non-center positions. The initialized weights are shown in Fig. \ref{fig_cc_initials}\subref{fig_cc_initial}. On this basis, the distance factor is further added. For example, for a $3\times 3$ convolution kernel, if the Euclidean distance is used to calculate the distance between the center position and other positions, there are only two distances of 1 and 1.414. The number of both distances is 4. Then the following equation can be constructed:
\begin{equation}
\begin{aligned}
\frac{x}{1} \times 4+\frac{x}{1.414} \times 4=1.
\end{aligned}
\end{equation}
Solving the equation to get $x$ equal to 0.146, then the weight of the position with a distance of 1 is 0.416, and the weight of the position with a distance of 1.414 is $x\div 1.414=0.104$. The overall weights are shown in Fig. \ref{fig_cc_initials}\subref{fig_cc_initial_distance}. We initialize the constrained convolution in this way.

Secondly, calculating the sum of the absolute values of the weights can avoid the result being negative or very small in absolute value. Thirdly, after the division operation, the small positive values and all negative values are set to 0.001 so that their effect is small and the values are not particularly small. Finally, the weight of the center position is not strictly equal to -1 but is equal to the negative sum of the weights of all non-center positions. With these constraints, the convolution can act as a high-pass filter and slow down fluctuations during training.

\subsection{Dual Branch Network and Edge Extraction Block}\label{dbn_eeb}
For the image manipulation detection task, the traces of image manipulation are very subtle. In addition to traces of manipulation regions, as described in Section \ref{relate_work}, the subtle differences between the manipulation edge and the non-manipulation region around it are also very important. To better capture this subtle information, features in CNN need to be maintained at a relatively high resolution. However, due to the local characteristics of convolution, it isn’t easy to obtain sufficient contextual information from high-resolution features.

To solve this problem, we draw on BiSeNet and DDRNet to design a dual branch network with Resnet-34 as the backbone. As shown in Fig. \ref{fig_model}, the high-resolution branch maintains the feature's height and width at 1/8 of the input image for more detailed information. The context branch changes the feature's height and width to 1/32 of the input image through multiple convolutions to obtain richer contextual information. Finally, the features of the two branches are fused to obtain information with sufficient context and details.

\begin{figure}[!t]
\centering
\includegraphics[width=2.8in]{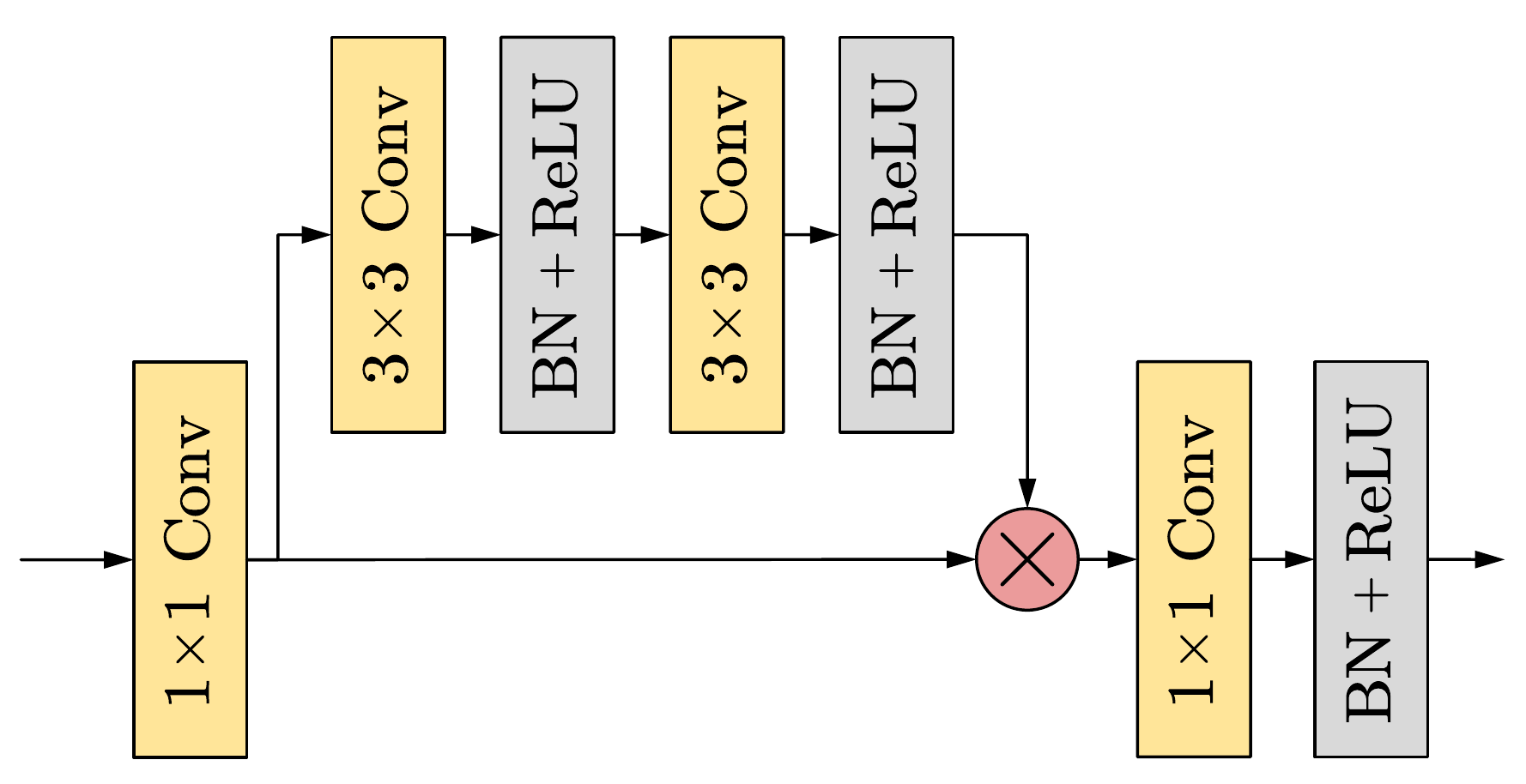}
\caption{Edge extraction block.}
\label{fig_eeb}
\end{figure}

Since each layer in the CNN learns different feature contents, we extract the manipulation edge based on the features output by each layer of the network. To extract edges better, we propose a specially designed edge extraction block (EEB). The flow of EEB is shown in Fig. \ref{fig_eeb}. To reduce the calculation and fully use the feature information, we first use $1\times 1$ convolution to reduce the number of features' channels to 1/4 of the original and then construct residual learning. Finally, $1\times 1$ convolution is used to reduce the number of channels to 1.

\subsection{Self-attention Mechanism with the Distance Factor}
The global correlations of pixels in features are essential for common CV tasks and image manipulation detection. Due to the locality limitation of convolution, it isn’t easy to obtain a complete global relationship. Therefore, similar to the self-attention mechanism in the NLP field, the non-local module is applied to CV tasks, which can break through the distance limitation of convolution and obtain the correlations between each pixel and other pixels.

\begin{figure}[!t]
\centering
\includegraphics[width=2.5in]{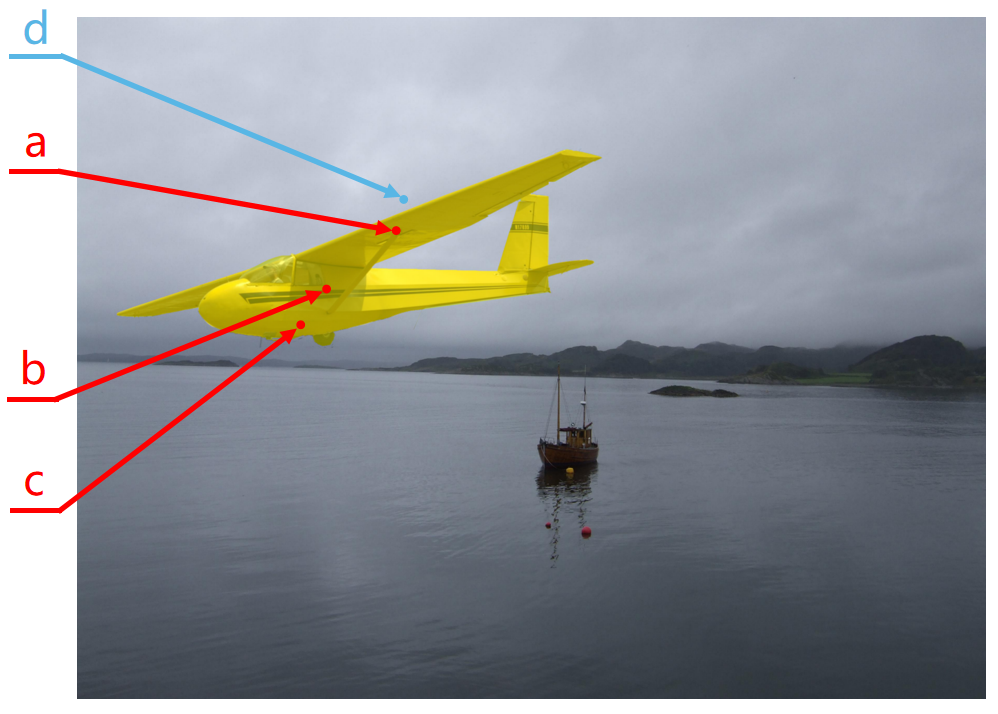}
\caption{Description of the correlations of points between different categories and different distances. The yellow part in the figure is class \emph{A}, and the rest is class \emph{B}. Points \emph{a}, \emph{b}, and \emph{c} both belong to class \emph{A}. Point \emph{d} belongs to class \emph{B}.}
\label{fig_distance_factor}
\end{figure}

However, ignoring the channel dimension, the image is 2-dimensional and contains much richer spatial information than 1-dimensional natural language. The influence of one pixel on other pixels in an image is closely related to distance. The closer two pixels are in the distance, the stronger their correlation is (pixels of the same class are positively correlated, and pixels of different classes are negatively correlated). As shown in Fig. \ref{fig_distance_factor}, it is assumed that the yellow part in the figure is class \emph{A}, and the rest is class \emph{B}. Points \emph{a}, \emph{b}, and \emph{c} belong to class \emph{A}, so they are positively correlated. But the distance from point \emph{a} to point \emph{b} is closer than the distance from point \emph{a} to point \emph{c}, so the correlation between point \emph{a} and point \emph{b} is stronger. Point a and d belong to different classes, so they correlate negatively. But the distance from point \emph{a} to point \emph{d} is closer than the distance from point \emph{a} to point \emph{b}, so the correlation between point \emph{a} and point \emph{d} is stronger.

\begin{figure}[!t]
\centering
\subfloat[\label{fig_distance_matrix_a}]{\includegraphics[width=0.5in]{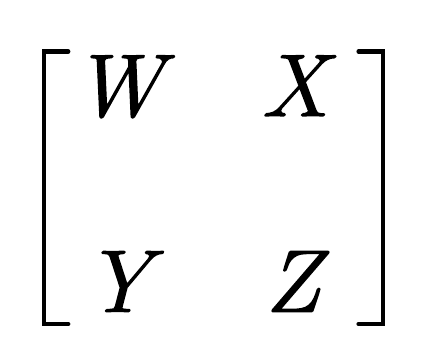}}\hspace{2mm}
\subfloat[\label{fig_distance_matrix_b}]{\includegraphics[width=1.35in]{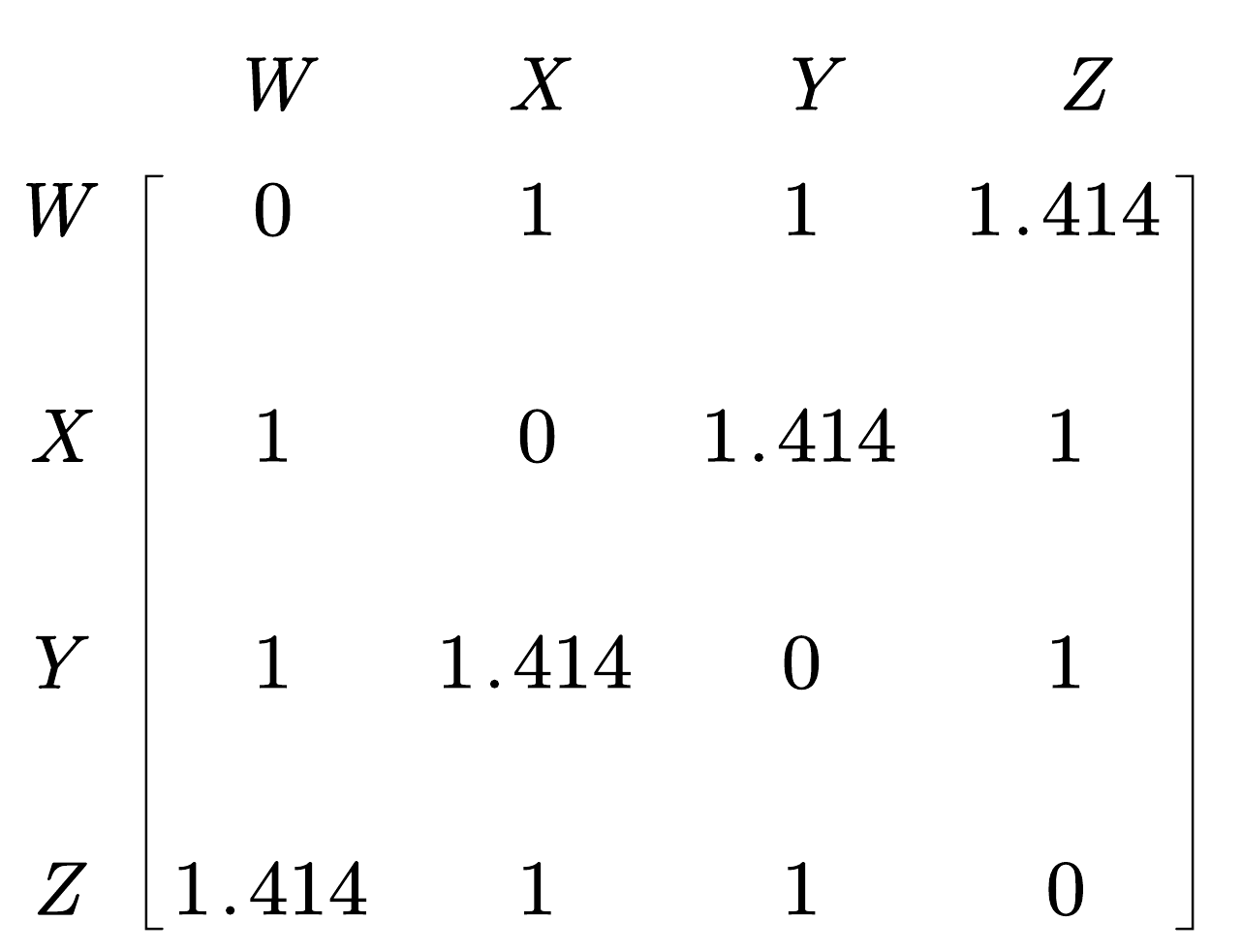}}\hspace{2mm}
\subfloat[\label{fig_distance_matrix_c}]{\includegraphics[width=1.35in]{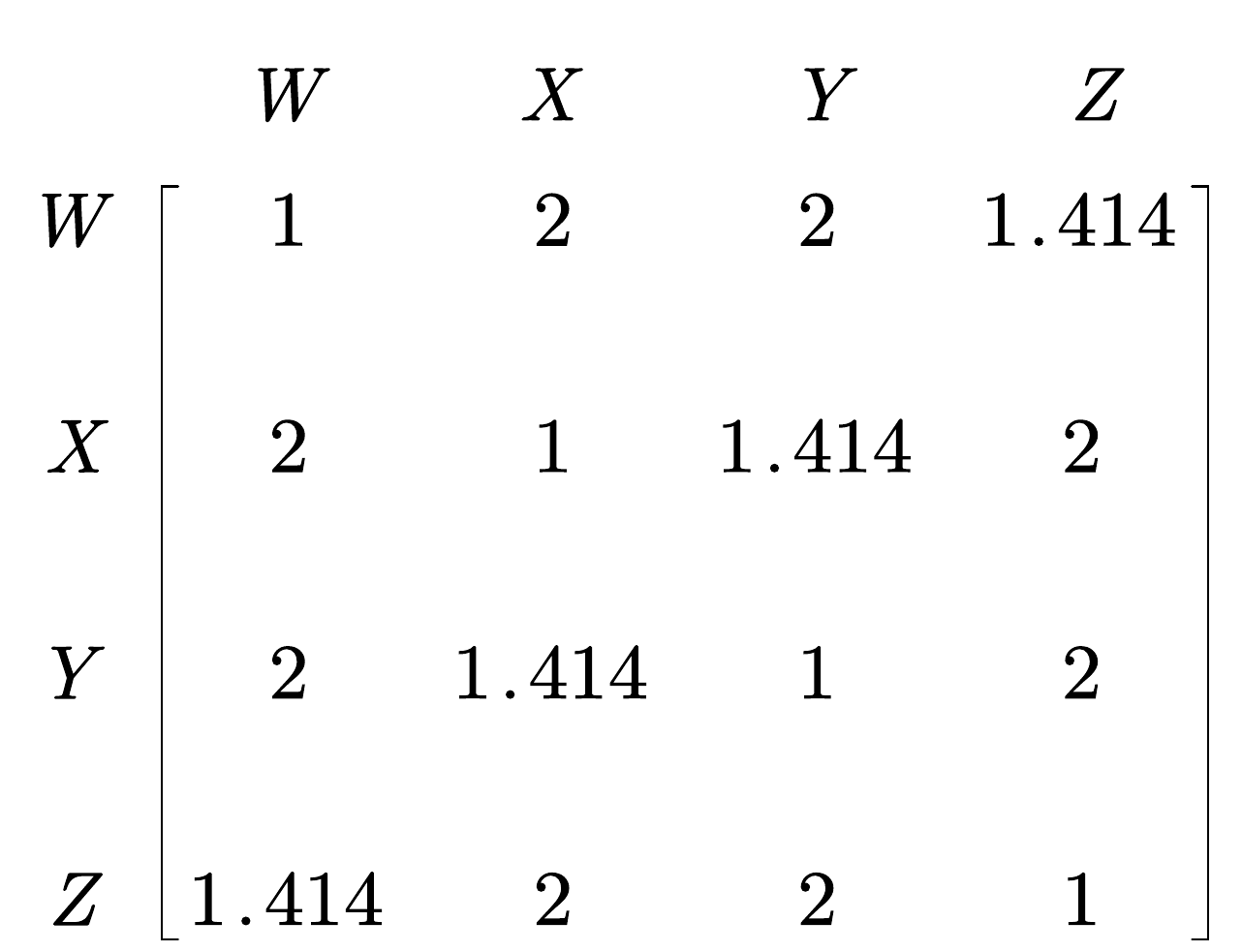}}
\caption{Distance matrix illustration. (a) An image with a height of 2 and a width of 2. (b) Distance matrix of the image. (c) Add 1 to the distance matrix value to avoid division by 0.}
\label{fig_distance_matrix}
\end{figure}

\begin{figure}[!t]
\centering
\includegraphics[width=3.2in]{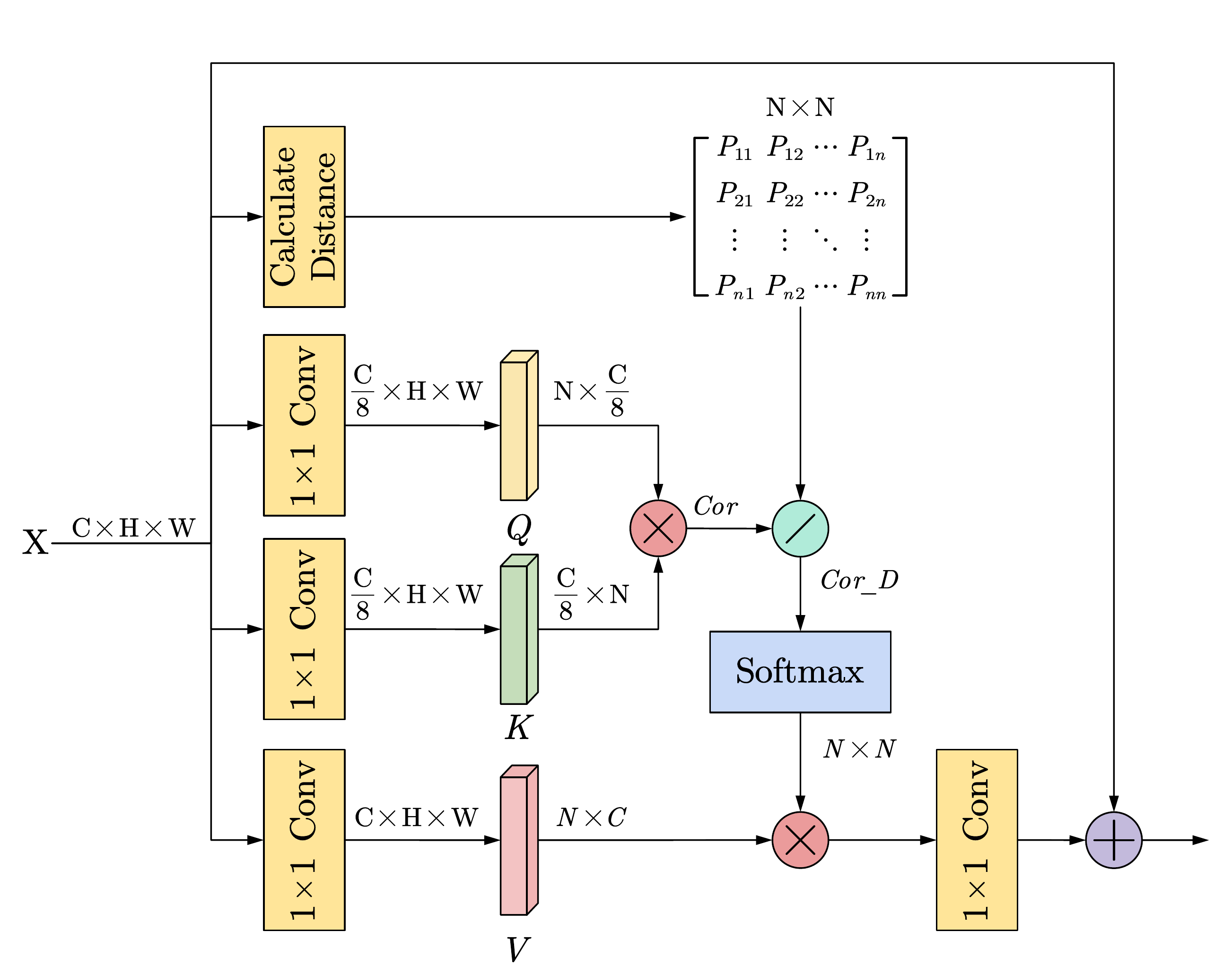}
\caption{The non-local module improved with the distance factor. C, H, and W represent the feature's dimension, height, and width. $\otimes$ denotes matrix multiplication, $\oplus$ denotes element-wise sum, and $\oslash$ denotes element-wise division. \emph{Cor} represents the underlying correlations among pixels, and \emph{Cor\_D} represents the distance-optimized correlations.}
\label{fig_non-local}
\end{figure}

The non-local module calculates the correlations among pixels across distances while ignoring the effect of distance factors on the correlation. To reflect the distance relationship between pixels, we use Euclidean distance to construct the distance matrix of each pixel to other pixels. As shown in Fig. \ref{fig_distance_matrix}\subref{fig_distance_matrix_a}, if we have an image with a height of 2 and a width of 2, then its corresponding distance matrix \emph{D} is shown in Fig. \ref{fig_distance_matrix}\subref{fig_distance_matrix_b}, with a size of $4\times 4$. The image has a total of 4 pixels. $D_{ij}$ in the distance matrix represents the distance from the \emph{i}-th pixel to the \emph{j}-th pixel, where \emph{i} is the row, and \emph{j} is the column. As shown in Fig. \ref{fig_non-local}, the matrix \emph{Cor} obtained by multiplying \emph{Q} and \emph{K} in the Non-local module represents the underlying correlations between each pixel and other pixels. We obtain the matrix \emph{Cor\_D} by element-wise division of the basic correlation matrix \emph{Cor} and the distance matrix \emph{D}. The matrix \emph{Cor\_D} represents the correlation after distance optimization. Then the final relationship among pixels is obtained through the Softmax operation. As shown in Fig. \ref{fig_distance_matrix}\subref{fig_distance_matrix_c}, to avoid the divisor being 0, we add 1 to the value of the distance matrix \emph{D}. 

\subsection{Manipulation Regions and Manipulation Edges Prediction}
\begin{figure}[!t]
\centering
\subfloat[]{\includegraphics[width=2.3in]{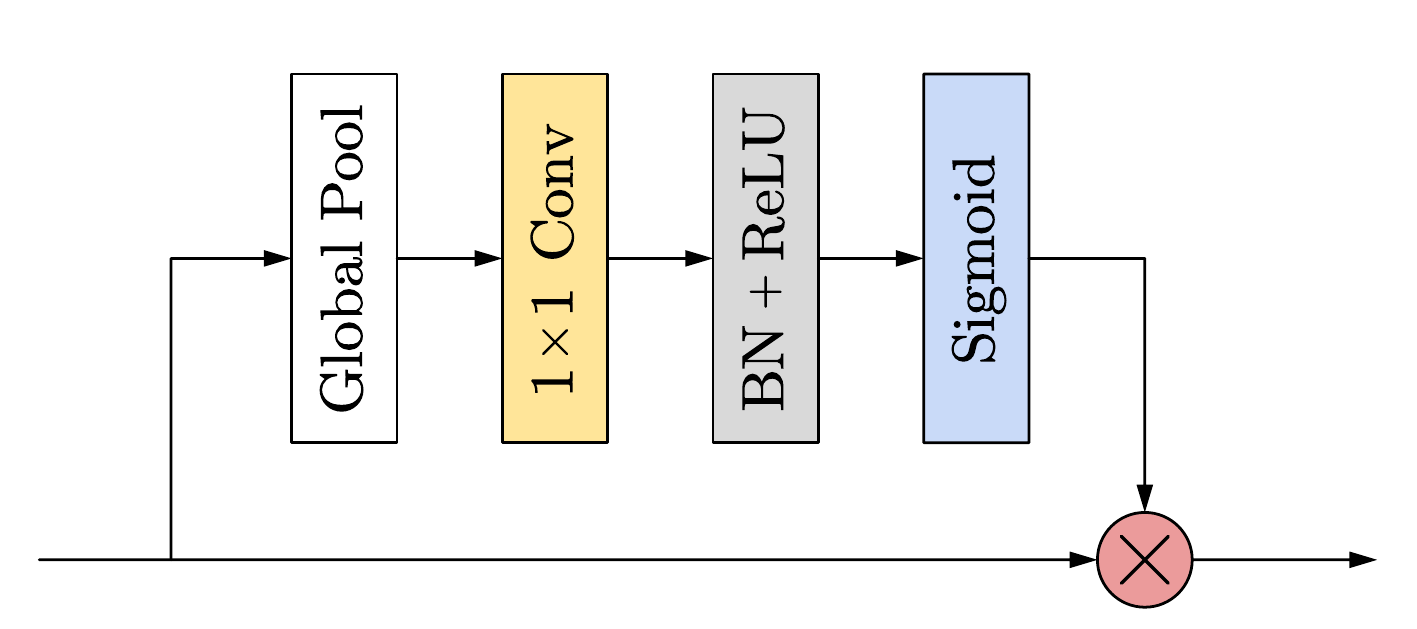}
\label{fig_bisenet_arm}}
\quad

\subfloat[]{\includegraphics[width=3in]{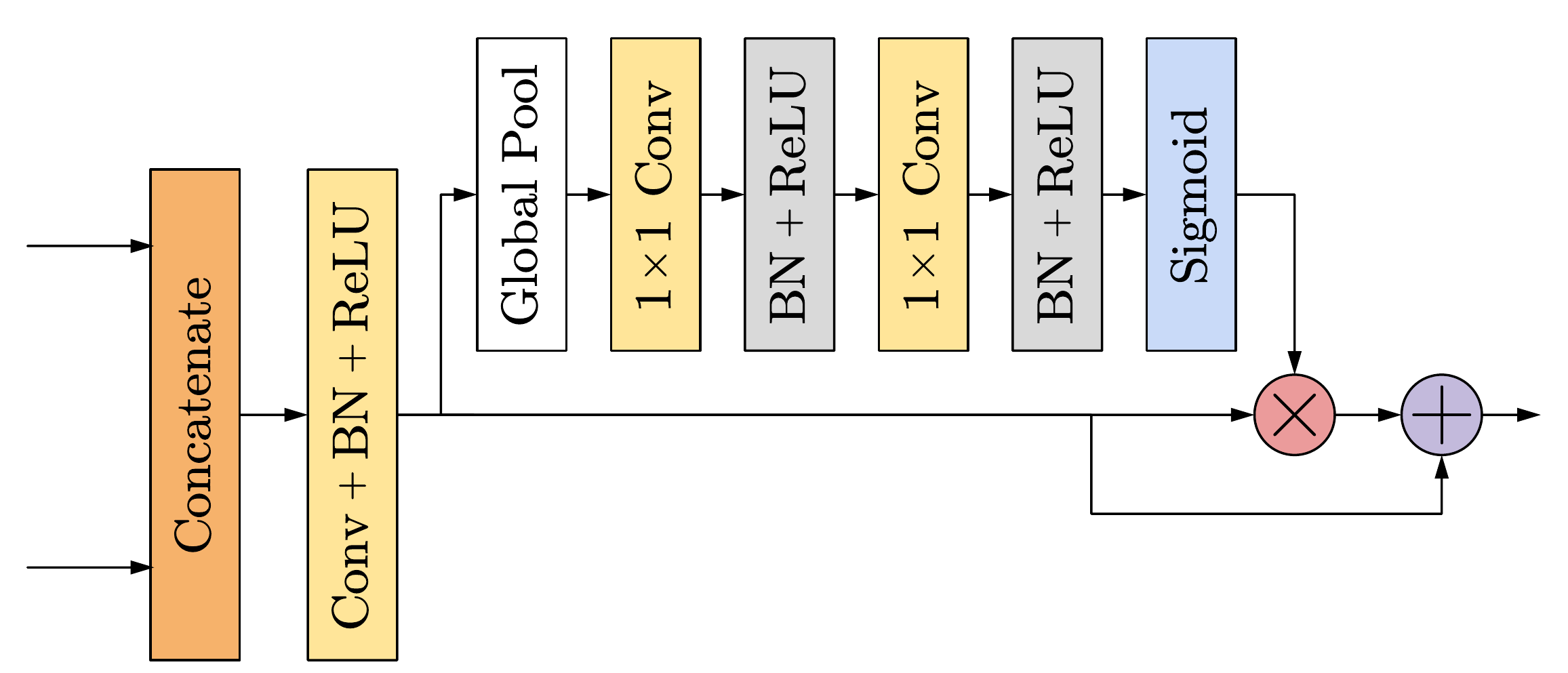}
\label{fig_bisenet_ffm}}
\caption{Two spatial attention modules. (a) Attention Refinement Module. (b) Feature Fusion Module}
\label{fig_bisenet}
\end{figure}

Since the features of the high-resolution branch and the context branch contain different granularities of information, we use BiSeNet's attention refinement module (ARM) and feature fusion module (FFM) to fuse them to predict the manipulation mask. ARM and FFM are shown in Fig. \ref{fig_bisenet}\subref{fig_bisenet_arm} and Fig. \ref{fig_bisenet}\subref{fig_bisenet_ffm}, respectively.

First, The ARM is used to optimize the features output by \emph{layer 3l} and \emph{layer 4l} in the channel dimension:
\begin{equation}
\left\{\begin{array}{l}
f_{3l\_A}=ARM(f_{3l}) \\
f_{4l\_A}=ARM(f_{4l})
\end{array}\right.,
\end{equation}
where $f_{3l}$ and $f_{4l}$ are the features output by \emph{layer 3l} and \emph{layer 4l}. $f_{3l_\_A}$ and $f_{4l_\_A}$ are ARM-optimized features. Then use the non-local module to perform self-attention calculation in the spatial dimension:
\begin{equation}
\left\{\begin{array}{l}
f_{3l\_N}=Nonlocal(f_{3l\_A}) \\
f_{4l\_N}=Nonlocal(f_{4l\_A})
\end{array}\right.,
\end{equation}
where $f_{3l\_N}$ and $f_{4l\_N}$ are features optimized by the non-local module. Then we perform the element-wise sum of the feature $f_{4h}$ output by \emph{layer 4h} with $f_{3l\_N}$ and $f_{4l\_N}$, respectively. Since they differ in resolution and dimensionality, they are processed using bilinear upsampling or $1\times 1$ convolution. Finally, use FFM to get the fused features:
\begin{equation}
\begin{aligned}
ff=FFM(&Up_{2\times 2}(f_{3l\_N})+Conv_{1\times 1}(f_{4h}),\\
&Up_{4\times 4}(f_{4l\_N})+f_{4h}),
\end{aligned}
\end{equation}
where $Up$ represents spatial upsampling, $Conv_{1\times 1}$ represents a $1\times 1$ convolution, and $ff$ represents the fused feature. The number of $ff$’s channels is 256.

$3\times 3$ convolution is performed after upsampling ff by a factor of 2. After repeating the above operation three times, the final prediction of manipulation regions is obtained. Three upsampling changes the resolution of ff to 1/4, 1/2, and 1/1 of the original input, respectively. That is, the resolution of the prediction result is the same as the original input. Three convolutions change the number of channels to 1/4, 1/16, and 1/256 of the initial $ff$. That is, the final number of channels in the prediction result is 1. These operations prevent the channel from decreasing too drastically while upsampling, thereby increasing the accuracy of the manipulation prediction.

We use the EEB in Section \ref{dbn_eeb} to separately predict the manipulation edges of the features output by the six layers in the model. Because the feature size in \emph{layer 1} production is 1/4 of the original input, we upsample the remaining five predicted manipulation edges by the same size. After concatenating the six prediction results in the channel dimension, a $3\times 3$ convolution is used to obtain the final predicted manipulation edge. The process is as follows:
\begin{equation}
\begin{aligned}
Pred_{edge}=Conv&_{3\times 3}(\\
Cat\{&edge_1, \space Up_{2\times 2}(edge_2),\\
&Up_{2\times 2}(edge_{3h}), \space Up_{4\times 4}(edge_{3l}),\\
&Up_{2\times 2}(edge_{4h}), \space Up_{8\times 8}(edge_{4l})\}),
\end{aligned}
\end{equation}
where $Cat$ represents the connection of the channel dimension, $edge_i$ represents the six edge prediction results, and $Conv_{3\times 3}$ represents the convolution of $3\times 3$. 

\begin{table}
\centering
\caption{The composition of manipulation datasets}
\label{tab_datasets_composition}
\begin{tabular}{l|c|c|c|c} 
\hline
Dataset  & Copy-move & Splicing & Removal & Total  \\ 
\hline
CASIAv1  & 459       & 461      & 0       & 920    \\ 
\hline
CASIAv2  & 3263      & 1843     & 0       & 5106   \\ 
\hline
COVERAGE & 93        & 0        & 0       & 93     \\ 
\hline
COLUMBIA & 0         & 180      & 0       & 180    \\ 
\hline
NIST16   & 68        & 288      & 208     & 564    \\
\hline
\end{tabular}
\end{table}

The manipulation regions of the image and the manipulation edges are small parts of the image. We use Dice Loss \cite{7785132} as the loss function to alleviate the problem of an unbalanced number of pixels in the manipulation and non-manipulation regions. The ground-truth of manipulation regions is processed by the dilation operation and the erosion operation to obtain $GT_D$ and $GT_E$, respectively. $GT_D-GT_E$ obtains the ground-truth of manipulation edges. The overall loss consists of the manipulation region prediction loss and the manipulation edge prediction loss. The calculation process is as follows:
\begin{equation}
Loss=\alpha \times loss_{region}+  (1 - \alpha) \times loss_{edge},
\end{equation}
where $loss_{region}$ represents the loss of the manipulation regions, $loss_{edge}$ represents the loss of the manipulation edges, and $\alpha$ is the weight.

\section{Experiment}
\subsection{Experimental Setup}
{\bf{Implementation Details.}} We implement our model using the Pytorch framework. The high-resolution branch and the context branch of the model are initialized with the weights of ResNet-34 pre-trained on ImageNet \cite{5206848}. The width and height of the input image are uniformly adjusted to 512. Image pixel values are divided by 255 for normalization. Standardize by subtracting the mean and dividing by the variance. The mean values of the three channels of BGR are 0.406, 0.456, and 0.485, respectively. The variances are 0.225, 0.224, and 0.229, respectively. Flipping and mirroring are used to perform simple data augment. The training batch size is set to 48. In the loss function, $\alpha$ is set to 0.3. We train the model for 12K stpdf. The learning rate is initially set to 0.01 and then reduced to 0.0075, 0.005, and 0.0025 after stpdf 5K, 7.5K, and 10K, respectively. All experiments are performed on a single NVIDIA Tesla V100 GPU with 32GB memory. The source code is available at: \url{https://github.com/kakashiz/NEDB-Net}.

{\bf{Evaluation Criteria.}} We evaluate the prediction results using pixel-level precision, recall, F1, and AUC. Because in data without GT, the most appropriate threshold cannot be predicted, we use the median value of 0.5 as the threshold to determine the positive and negative classes.

\begin{table*}
\centering
\caption{Performance of pixel-level manipulation detection}
\label{tab_cp_f1}
\begin{tabular}{l|c|c|c|c|c} 
\hline
           & CASIAv1          & COVERAGE         & COLUMBIA         & NIST16           & Mean              \\ 
\hline
FCN        & 0.441            & 0.199            & 0.223            & 0.167            & 0.258             \\ 
\hline
HP-FCN     & 0.154            & 0.003            & 0.067            & 0.121            & 0.086             \\ 
\hline
Mantra-Net & 0.155            & 0.286            & 0.364            & 0.000            & 0.201             \\ 
\hline
CR-CNN     & 0.405            & 0.291            & 0.436            & 0.238            & 0.343             \\ 
\hline
GSR-Net    & 0.387            & 0.285            & 0.613            & 0.283            & 0.392             \\ 
\hline
MVSS-Net   & 0.452            & 0.453            & 0.638            & 0.292            & 0.459             \\ 
\hline
D-FCN (pre-train)      & 0.007                & 0.194                & 0.376                & \textbf{ 0.402 }               & 0.245                 \\ 
\hline
D-FCN (re-train)      & 0.331                & 0.260               & 0.233                & 0.136               & 0.240                \\ 
\hline
Ours       & \textbf{ 0.516 } & \textbf{ 0.461 } & \textbf{ 0.756 } & 0.288            & \textbf{ 0.505 }  \\
\hline
\end{tabular}
\end{table*}

\subsection{Comparison with the State of the art}\label{sub_section_cp}
{\bf{Datasets.}} We select CASIAv1 \cite{dong2010casia}, CASIAv2 \cite{dong2013casia}, COVERAGE \cite{wen2016coverage}, COLUMBIA \cite{hsu2006columbia}, NIST16 \cite{guan2019mfc} these publicly available image manipulation datasets as the experimental datasets. The composition of the manipulation image types of datasets is shown in Table \ref{tab_datasets_composition}. It is worth noting that some of the manipulation images lack the corresponding ground-truth or do not match the shape of the ground-truth. We only count the correct manipulation images here.

{\bf{Evaluation method.}} When comparing effects in many studies, the model will be trained on other large datasets and then tested on the above datasets. Or select a portion of each dataset for fine-tuning and another portion for testing. We do not think this comparison is particularly plausible because it does not adequately demonstrate the model's generality against unknown data. So, we adopt the same evaluation method as \cite{Chen_2021_ICCV}, let the model train only on the CASIAv2 dataset, and then directly test it on the remaining dataset. This approach directly reflects whether the model has learned how to detect manipulation rather than just fitting the dataset.

{\bf{Baselines.}} The baseline models we chose are as follows: Fully Convolutional Networks (FCN) \cite{Long_2015_CVPR}, High-Pass Fully Convolutional Network (HP-FCN) \cite{Li_2019_ICCV}, Manipulation Tracing Network (Mantra-Net) \cite{Wu_2019_CVPR}, Constrained R-CNN (CR-CNN) \cite{9102825}, Generate, Segment, Refine Network (GSR-Net) \cite{zhou2020generate}, Multi-View Multi-Scale Supervision Network (MVSS-Net) \cite{Chen_2021_ICCV}, and Dense Fully Convolutional Network (D-FCN) \cite{9393396}. Among them, HP-FCN and ManTra-Net directly use the models provided by the authors due to the lack of training codes and private datasets. The authors provide the weight trained on the private manipulation dataset and 10\% of the NIST16 dataset. We train on the CASIAv2 dataset based on this pretrained weight and report two experimental results, where D-FCN (pre-train) means directly using the author's pre-trained weights for testing, and D-FCN  (re-train) means using our retrained model for testing. Other methods either follow the same evaluation protocol or retrain on the CASIAv2 dataset. The experimental results are shown in Table \ref{tab_cp_f1} (part of the experimental results are obtained from \cite{Chen_2021_ICCV}).

{\bf{Manipulation detection effect comparison.}} As can be seen from Table \ref{tab_cp_f1}, our model outperforms other models on other datasets, except that it is worse than D-FCN (pre-train) and MVSS-Net on the NIST16 dataset. Especially the results on COLUMBIA and CASIAv1 are much ahead of other models. The experimental results verify the detection effect of our model and show that our model can also achieve good results on unknown data sets.

\begin{figure}[!t]
\centering
\subfloat[]{\includegraphics[width=3in]{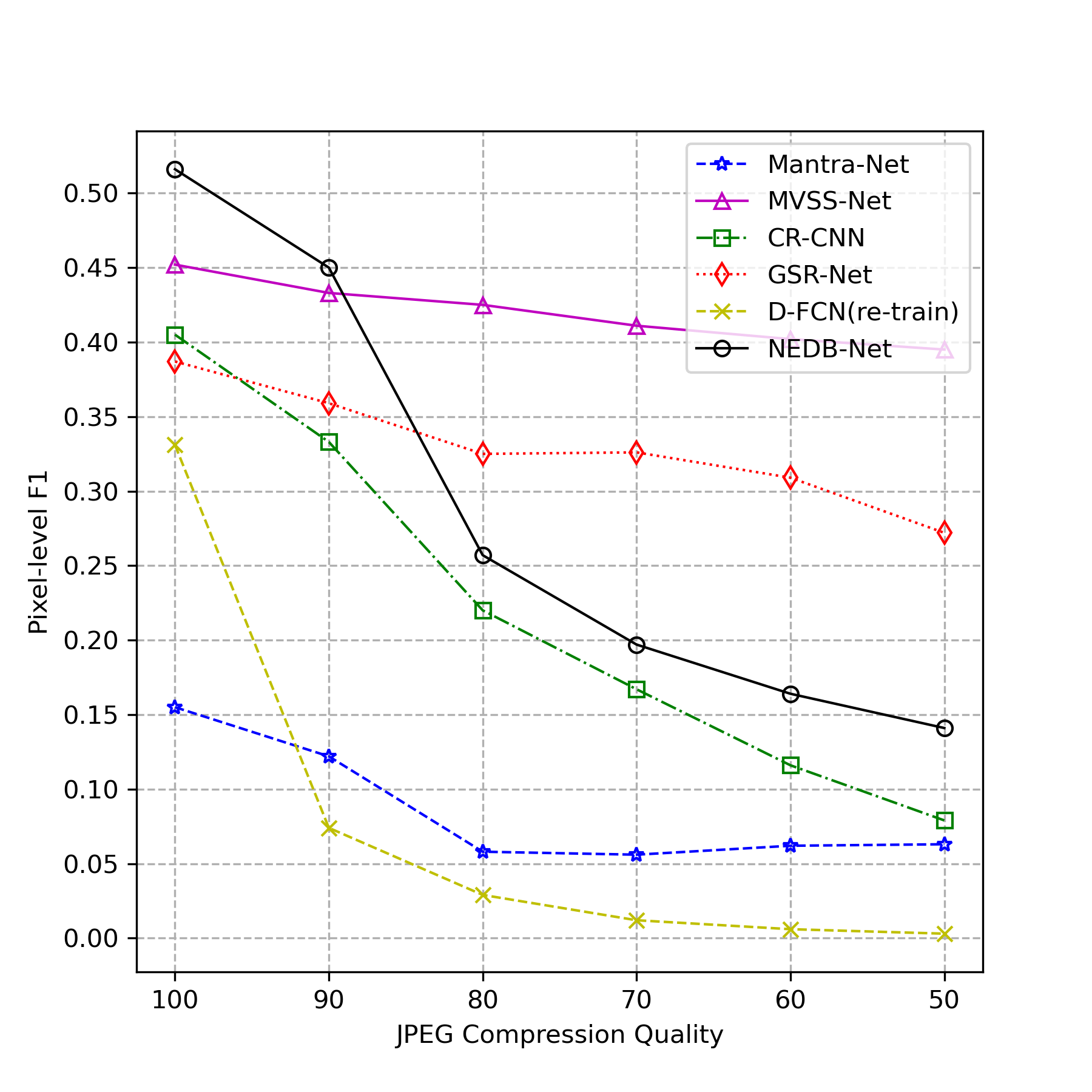}
\label{robust_jpeg_compression}}
\quad
\subfloat[]{\includegraphics[width=3in]{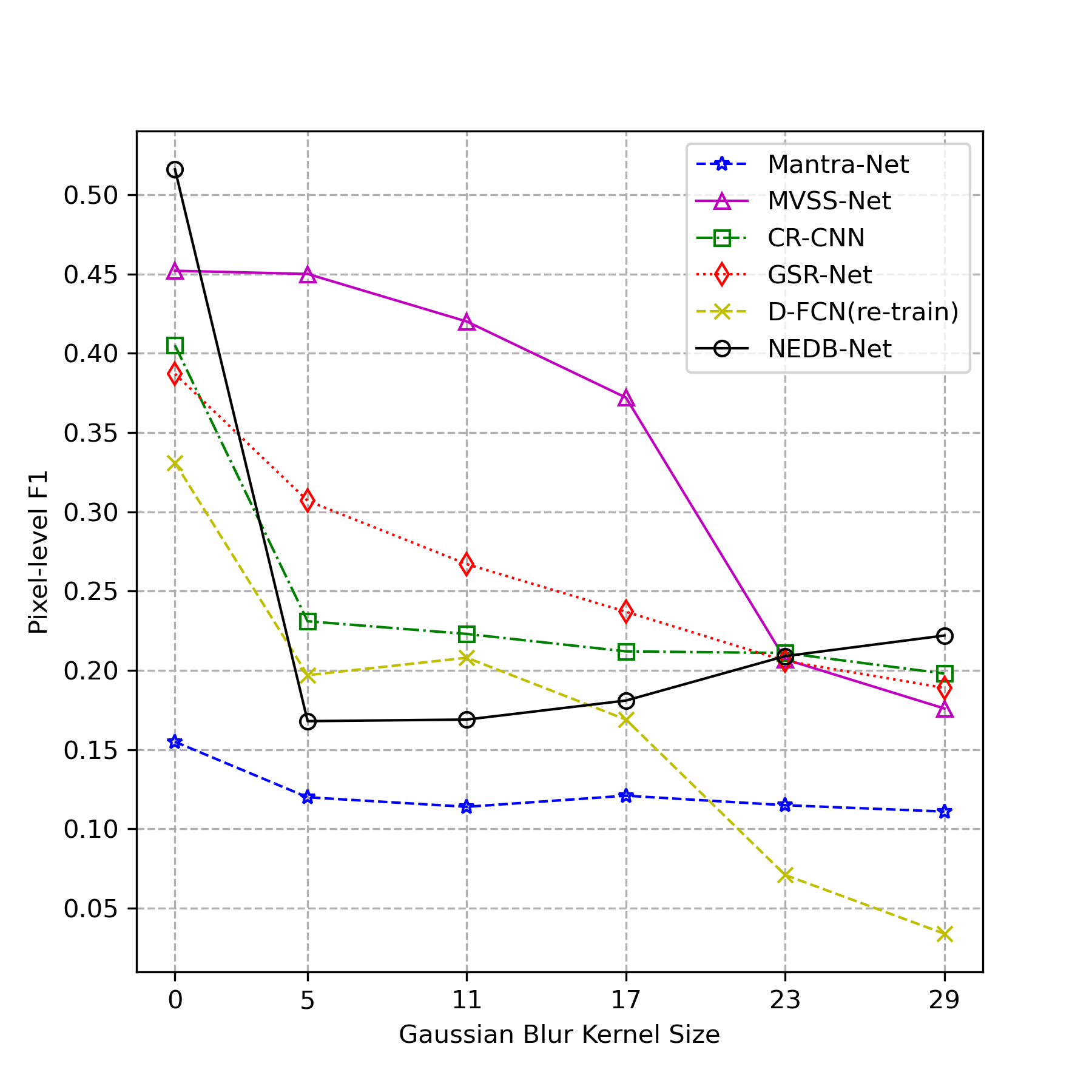}
\label{robust_gaussian_blur}}
\caption{Robustness evaluation on CASIAv1. (a) Performance curves w.r.t. JPEG compression. (b) Performance curves w.t.r. Gaussian Blurs.}
\label{fig_robust}
\end{figure}

{\bf{Robustness evaluation.}} To compare the robustness of the models, we follow the same approach as \cite{Chen_2021_ICCV}, applying different levels of JPEG compression and Gaussian blur to CASIAv1, respectively. The experimental results are shown in Fig. \ref{fig_robust}. From this, we can see that our model is not ideal in terms of robustness performance. As the interference to the original image gradually increases, the detection effects of CR-CNN and our model decay more than the detection effects of other models. For this case, one possible reason is that both CR-CNN and our model only use the noise image as the model's input. Compression and blurring damage the noise information more than semantic information. On the other hand, if the entire image is compressed or blurred, can the whole image be considered a manipulation region? If so, the above experimental results are not accurate. However, many studies' robustness experiments use post-processing on the whole image. We think there should be a better way for ablation experiments.

\subsection{Ablation Study}
\label{subsec_ablation}
To better illustrate the effect of each module in our model, we gradually add each component on ResNet-34 for training and compare the results. Since the number of images in the datasets mentioned in Section \ref{sub_section_cp} is small, to make the model fully trained, we use the COCO \cite{lin2014microsoft} dataset to generate a large number of simple manipulation images as the dataset for ablation experiments. The COCO dataset contains a large number of images and annotations of objects in the images. Based on these object annotations, manipulation images can be easily created. The generation process of the manipulation image is as follows:
\begin{enumerate}
\item{Randomly select one image (copy-move type) or two images (splicing type) from the dataset as the source and destination images. The source and destination images are the same if only one image is selected.}
\item{Randomly select an object from the source image, and perform operations such as rotation and scaling on the object.}
\item{Paste the object randomly to a certain position of the destination image.}
\item{Perform post-processing operations such as blurring.}
\end{enumerate}

A total of 76K synthetic manipulation images were generated using the above method. We train models using these manipulation images and test them on the COLUMBIA dataset and the CASIAv1 dataset. The models for the ablation experiment are as follows:
\begin{itemize}
\item{\emph{SB}: Use ResNet-34 as the base model.}
\item{\emph{DB}: Use ResNet-34 to build a dual-branch network and add the ARM module and FFM module for features fusion.}
\item{\emph{DB+origin-CC}: Add the original constrained convolution on the dual-branch network.}
\item{\emph{DB+CC}: Add the improved constrained convolution on the dual-branch network.}
\item{\emph{DB+CC+Edge}: Add the improved constrained convolution and edge detection on the dual-branch network.}
\item{\emph{DB+CC+Edge+NL}: Add the improved constrained convolution, edge detection, and Non-local module on the dual branch network.}
\item{\emph{DB+CC+Edge+NL-D}: Add the improved constrained convolution, edge detection, and Non-local module with distance factor on the dual-branch network, i.e. the final model.}
\end{itemize}

The results of the ablation experiments are shown in Table \ref{tab_ab}. Comparing \emph{SB} and \emph{DB} shows that the dual-branch network can capture more detailed information that is beneficial for manipulation detection. The comparison of the results of \emph{DB}, \emph{DB+origin-CC}, and \emph{DB+CC} confirms the problem of the original constrained convolution we mentioned in Section \ref{subsec_cc} and illustrates the effect of our improved constrained convolution. From the comparison between \emph{DB+CC} and \emph{DB+CC+Edge}, we can find that manipulation edge detection is very helpful to the overall detection effect. The results of \emph{DB+CC+Edge}, \emph{DB+CC+Edge+NL}, and \emph{DB+CC+Edge+NL-D} show that the global correlations computed by the self-attention mechanism are beneficial to distinguishing manipulation and non-manipulation pixels, and the self-attention mechanism after adding the distance factor works better.

\begin{table*}
\centering
\caption{Ablation experiment results}
\label{tab_ab}
\begin{tabular}{l|c|c|c|c|c|c|c|c} 
\hline
\multirow{2}{*}{Models} & \multicolumn{4}{c|}{COLUMBIA}                                     & \multicolumn{4}{c}{COVERAGE}                                      \\ 
\cline{2-9}
                        & Precision      & Recall         & F1             & AUC            & Precision      & Recall         & F1             & AUC             \\ 
\hline
SB                      & 0.821          & 0.655          & 0.719          & 0.808          & 0.610          & 0.438          & 0.474          & 0.714           \\ 
\hline
DB                      & 0.814          & 0.719          & 0.750          & 0.831          & 0.632          & 0.498          & 0.505          & 0.739           \\ 
\hline
DB+origin-CC            & 0.766          & 0.717          & 0.741          & 0.822          & 0.651          & 0.533          & 0.553          & 0.752           \\ 
\hline
DB+CC                   & 0.803          & 0.765          & 0.771          & 0.845          & 0.632          & 0.591          & 0.568          & 0.771           \\ 
\hline
DB+CC+Edge              & 0.831          & 0.755          & 0.784          & 0.851          & 0.665          & 0.613          & 0.602          & 0.796           \\ 
\hline
DB+CC+Edge+NL           & 0.801          & \textbf{0.809} & 0.794          & 0.862          & \textbf{0.696} & 0.718          & 0.676          & 0.831           \\ 
\hline
DB+CC+Edge+NL-D         & \textbf{0.848} & 0.800          & \textbf{0.815} & \textbf{0.872} & 0.682          & \textbf{0.769} & \textbf{0.688} & \textbf{0.851}  \\
\hline
\end{tabular}
\end{table*}

\begin{table*}
\centering
\caption{Experimental results of different initialization methods for the constrained convolution}
\label{tab_cc_initials}
\begin{tabular}{l|c|c|c|c|c|c|c|c} 
\hline
\multirow{2}{*}{Initialization
  method} & \multicolumn{4}{c|}{COLUMBIA}                                     & \multicolumn{4}{c}{COVERAGE}                                        \\ 
\cline{2-9}
                                         & \multicolumn{1}{c|}{Precision} & Recall         & F1             & AUC            & Precision      & Recall         & F1             & AUC             \\ 
\hline
Random1                                  & 0.830                          & 0.741          & 0.769          & 0.848          & 0.717          & 0.587          & 0.604          & 0.785           \\ 
\hline
Random2                                  & 0.830                          & 0.766          & 0.786          & 0.853          & \textbf{0.721} & 0.681          & 0.645          & 0.822           \\ 
\hline
Random3                                  & 0.816                          & 0.795          & 0.798          & 0.858          & 0.687          & 0.703          & 0.640          & 0.824           \\ 
\hline
Random-sum1                              & 0.807                          & \textbf{0.809} & 0.799          & 0.856          & 0.701          & 0.757          & 0.688          & 0.855           \\ 
\hline
Random-sum2                              & 0.817                          & 0.760          & 0.781          & 0.849          & 0.693          & 0.566          & 0.576          & 0.773           \\ 
\hline
Random-sum3                              & 0.824                          & 0.797          & 0.799          & 0.859          & 0.707          & 0.636          & 0.636          & 0.807           \\ 
\hline
Laplace-like                             & 0.807                          & 0.809          & 0.798          & 0.856          & 0.701          & 0.757          & 0.688          & 0.849           \\ 
\hline
Laplace-like-D                           & \textbf{0.848}                 & 0.800          & \textbf{0.815} & \textbf{0.872} & 0.682          & \textbf{0.769} & \textbf{0.688} & \textbf{0.851}  \\
\hline
\end{tabular}
\end{table*}

\subsection{Constrained convolution initialization}
Since our model uses the noise image extracted by the constrained convolution instead of the original image as the manipulation detection cue, the initialization method of the constrained convolution is very important for the model network. The initialization methods for comparison are as follows: 

\begin{itemize}
\item{\emph{Random}: The weight of the center position is -1, and the weight of other positions is randomly initialized to a value between 0-1.}
\item{\emph{Random-Sum}: The center position weight is -1. The weight of other positions is randomly initialized to a value between 0-1, and the sum of the weights of the non-central positions is 1.}
\item{\emph{Laplace-like}: The initialization process is described in Section \ref{subsec_cc} and Fig. \ref{fig_cc_initials}\subref{fig_cc_initial}.}
\item{\emph{Laplace-like-D}: The initialization process is described in Section \ref{subsec_cc} and Fig. \ref{fig_cc_initials}\subref{fig_cc_initial_distance}.}
\end{itemize}

The training and test sets of the experiments are the same as in Section \ref{subsec_ablation}. Since \emph{Random} and \emph{Random-Sum} are initialized randomly, we retrain them three times. The experimental results are shown in Table \ref{tab_cc_initials}. It can be seen that the results of \emph{Random} and \emph{Random-Sum} are both unstable due to random initialization. But the initialization of \emph{Random-Sum} is similar to a high-pass filter, so the overall effect is better than \emph{Random}. \emph{Laplace-like} and \emph{Laplace-like-D} perform better than the two random initialization methods. And \emph{Laplace-like-D} has a better overall effect because it can better reflect the correlation between pixels according to the distance in the initialization process.

\subsection{Kernel size of constrained convolution}
Different sizes of convolution kernels greatly influence the effect of constrained convolution. If the convolution kernel is too small, its receptive field is too small to capture complete information. If the convolution kernel is too large, it will also capture a lot of irrelevant information while affecting the calculation speed. So, we gradually increase the convolution size of the constrained convolution kernel from $3\times 3$ to $11\times 11$ for experiments. The training and test sets of the experiments are the same as in Section \ref{subsec_ablation}. Experimental results are shown in Table \ref{tab_cc_kernel_size}. From the experimental results, we can see that when the convolution kernel size is $5\times 5$, the overall detection effect is better. Although the effect is also good when the convolution kernel size is $
11\times 11$, a too-large convolution kernel will affect the calculation speed.

\begin{table*}
\centering
\caption{Experimental results of different kernel sizes for the constrained convolution}
\label{tab_cc_kernel_size}
\begin{tabular}{l|c|c|c|c|c|c|c|c} 
\hline
\multirow{2}{*}{Kernal
  size} & \multicolumn{4}{c|}{COLUMBIA}                                     & \multicolumn{4}{c}{COVERAGE}                                         \\ 
\cline{2-9}
                               & Precision      & Recall         & F1             & AUC            & Precision       & Recall         & F1             & AUC             \\ 
\hline
3x3                            & 0.810          & 0.722          & 0.751          & 0.835          & 0.667           & 0.589          & 0.583          & 0.786           \\ 
\hline
5x5                            & \textbf{0.848} & 0.800          & \textbf{0.815} & \textbf{0.872} & 0.682           & 0.769          & 0.688          & 0.851           \\ 
\hline
7x7                            & 0.823          & 0.775          & 0.791          & 0.852          & 0.690           & 0.626          & 0.620          & 0.795           \\ 
\hline
9x9                            & 0.836          & 0.811          & 0.813          & 0.873          & 0.693           & 0.680          & 0.648          & 0.818           \\ 
\hline
11x11                          & 0.773          & \textbf{0.853} & 0.794          & 0.862          & \textbf{0.725 } & \textbf{0.778} & \textbf{0.694} & \textbf{0.857}  \\
\hline
\end{tabular}
\end{table*}

Since the number of channels of the input image is three, to keep the number of channels of the noise image consistent, the size of the constrained convolution in the batch dimension is three. Different sizes of convolution kernels capture different information, but it is questionable whether setting the convolution kernels of different batches to different sizes can get better results. To verify this question, we conduct experiments with the following settings:
\begin{itemize}
\item{\emph{3, 5, 7}: The sizes of three batches of convolution kernels are set to $3\times 3$, $5\times 5$, and $7\times 7$, respectively.}
\item{\emph{5, 7, 9}: The sizes of three batches of convolution kernels are set to $5\times 5$, $7\times 7$, and $9\times 9$, respectively.}
\item{\emph{5, 5, 5}: The sizes of three batches of convolution kernels are all set to $5\times 5$.}
\end{itemize}

The experimental results are shown in Table \ref{tab_cc_kernel_combine}. Contrary to our assumption, combining convolution kernels of different sizes does not improve the detection effect but leads to poor detection results. One possible reason for the poor results is that the noise features extracted by different convolution kernels are quite different. These differences are not conducive to subsequent predictions.

\begin{table*}
\centering
\caption{Experimental results of combining different sizes of convolution kernels}
\label{tab_cc_kernel_combine}
\begin{tabular}{l|c|c|c|c|c|c|c|c} 
\hline
\multirow{2}{*}{Kernal
  size} & \multicolumn{4}{c|}{COLUMBIA}                                     & \multicolumn{4}{c}{COVERAGE}                                       \\ 
\cline{2-9}
                               & Precision      & Recall         & F1             & AUC            & Precision      & Recall         & F1             & AUC             \\ 
\hline
3, 5,
  7                      & 0.753          & 0.661          & 0.682          & 0.795          & 0.494          & 0.526          & 0.462          & 0.725           \\ 
\hline
5, 7,
  9                      & 0.765          & 0.656          & 0.684          & 0.791          & 0.502          & 0.492          & 0.451          & 0.713           \\ 
\hline
5, 5,
  5                      & \textbf{0.848} & \textbf{0.800} & \textbf{0.815} & \textbf{0.872} & \textbf{0.682} & \textbf{0.769} & \textbf{0.688} & \textbf{0.851}  \\
\hline
\end{tabular}
\end{table*}

\begin{table*}
\centering
\caption{Influence of manipulation edge ground-truth generated by different methods on manipulation detection; \emph{ELLIPSE} means the kernel is elliptical; \emph{RECT} means the kernel is rectangular; \emph{CROSS} means the kernel is cross-shaped}
\label{tab_edge_kernel}
\begin{tabular}{l|c|c|c|c|c|c|c|c} 
\hline
\multirow{2}{*}{Edge kernel
} & \multicolumn{4}{c|}{COLUMBIA}                                     & \multicolumn{4}{c}{COVERAGE}                                         \\ 
\cline{2-9}
                             & Precision      & Recall         & F1             & AUC            & Precision      & Recall         & F1             & AUC             \\ 
\hline
ELLIPSE\_$3\times 3$                 & 0.845          & 0.777          & 0.799          & 0.859          & \textbf{0.740} & 0.635          & 0.645          & 0.808           \\ 
\hline
ELLIPSE\_$5\times 5$                 & 0.848          & 0.800          & \textbf{0.815} & \textbf{0.872} & 0.682          & \textbf{0.769} & \textbf{0.688} & \textbf{0.851}  \\ 
\hline
ELLIPSE\_$7\times 7$                 & 0.830          & 0.798          & 0.803          & 0.867          & 0.707          & 0.661          & 0.638          & 0.816           \\ 
\hline
ELLIPSE\_$9\times 9$                 & 0.836          & 0.797          & 0.807          & 0.864          & 0.681          & 0.665          & 0.645          & 0.813           \\ 
\hline
RECT\_$3\times 3$                   & 0.827          & \textbf{0.808} & 0.805          & 0.868          & 0.689          & 0.693          & 0.660          & 0.827           \\ 
\hline
RECT \_$5\times 5$                   & 0.831          & 0.769          & 0.791          & 0.857          & 0.727          & 0.707          & 0.675          & 0.837           \\ 
\hline
RECT \_$7\times 7$                   & 0.809          & 0.787          & 0.787          & 0.854          & 0.682          & 0.693          & 0.647          & 0.825           \\ 
\hline
RECT \_$9\times 9$                   & 0.837          & 0.784          & 0.801          & 0.863          & 0.669          & 0.624          & 0.598          & 0.797           \\ 
\hline
CROSS\_$3\times 3$                   & 0.834          & 0.793          & 0.801          & 0.865          & 0.701          & 0.640          & 0.625          & 0.804           \\ 
\hline
CROSS \_$5\times 5$                  & 0.847          & 0.782          & 0.804          & 0.864          & 0.668          & 0.634          & 0.615          & 0.802           \\ 
\hline
CROSS \_$7\times 7$                  & 0.798          & 0.777          & 0.775          & 0.846          & 0.707          & 0.716          & 0.670          & 0.840           \\ 
\hline
CROSS \_$9\times 9$                  & \textbf{0.851} & 0.767          & 0.796          & 0.862          & 0.661          & 0.611          & 0.607          & 0.796           \\
\hline
\end{tabular}
\end{table*}

\subsection{Generation method of manipulation edge ground-truth}
As mentioned earlier, the manipulation edge is a vital clue. The ground-truth of manipulation regions is processed by OpenCV's dilation and erosion operations to obtain $GT_D$ and $GT_E$. $GT_D-GT_E$ gets the ground-truth of manipulation edges. In OpenCV's dilation and erosion operations, the shape and size of the kernel can be set. The shape of the kernel can be oval, rectangular, or cross. The width of manipulation edges in the ground-truth can be adjusted by setting the size of the kernel. If the edge is too narrow, it is tough for training and prediction. If the edge is too wide, it will cause too many irrelevant pixels to affect the prediction effect.

So, we use kernels of different shapes and sizes to generate the ground-truth of manipulation edges for experiments. The training and test sets of the experiments are the same as in Section \ref{subsec_ablation}. Experimental results are shown in Table \ref{tab_edge_kernel}. From this table, we can see that the average effect of elliptical and rectangular kernels is better than that of cross-shaped kernels, and the elliptical kernels perform best. One possible reason is that the number of pixels involved in the cross-shaped kernel operation is too small, while the rectangular kernel operation involves more irrelevant pixels. In addition, the detection performance does not continually improve as the kernel size increases. That is, the width of the ground-truth of edges should be moderate. In our experiments, the edges produced by the $5\times 5$ kernel performed best for manipulation detection.

\section{Conclusion}
In this paper, we propose an image manipulation detection model based on image noise and manipulation edge. The improved constrained convolution can better extract the noise information of the image while addressing the training stability. The Non-local module ignores the distance relationship among pixels while computing the correlations among pixels spanning distances. We add a distance factor to it to better capture the global information among pixels. The manipulation edge detection task constructed based on the features of the high-resolution branch and the context branch greatly improves the overall manipulation detection effect. Experiments on public manipulation datasets and synthetic manipulation datasets demonstrate the state-of-the-art detection performance of our model.

There are still some problems to be solved here. Firstly, the noise information of the image is not robust to JPEG compression and Gaussian blur. In addition to noise, how can factors such as illumination and chromaticity related to manipulation inconsistency be better applied in neural networks? Secondly, there is disagreement in the identification of large manipulation regions. If 90\% of an image is spliced, is it correct for the model to identify the original 10\% as manipulation? In the future we will further optimize the structure of the model.

\section*{Acknowledgments}
The authors would like to thank the Jiutian deep learning platform of China Mobile. This work used the platform for project construction and model training.

\bibliography{reference}
\bibliographystyle{IEEEtran.bst}

\vfill

\end{document}